\PassOptionsToPackage{table}{xcolor} 
\documentclass[runningheads]{llncs}

\usepackage{eccv}
\usepackage{tcolorbox}
\usepackage{soul}
\usepackage{wrapfig}
\usepackage{eccvabbrv}
\usepackage{amsmath}
\usepackage{amsfonts} 
\usepackage{amssymb} 
\usepackage{multirow}
\usepackage{threeparttable}
\usepackage{booktabs}
\usepackage{graphicx}
\usepackage{soul}

\usepackage{xcolor} 
\usepackage{pifont}
\usepackage{float}
\usepackage[accsupp]{axessibility}
\newcommand{\xmark}{\ding{55}} 
\definecolor{DeepRed}{RGB}{160, 0, 0}
\definecolor{LogicGray}{gray}{0.82}
\sethlcolor{LogicGray}
\newcommand{\target}[1]{\textbf{\textcolor{DeepRed}{#1}}}
\newcommand{\spatial}[1]{\hl{#1}}

\usepackage{hyperref}

\usepackage{orcidlink}

\begin{document}

\title{CROSS: Cascaded Distillation and Dual-Constraint Grounding for Remote Sensing Referring Segmentation} 

\titlerunning{CROSS}

\newcommand{\equalcontrib}{\textsuperscript{*}}
\newcommand{\corrauth}{\textsuperscript{\ensuremath{\dagger}}}

\author{
Tingzhang Luo\equalcontrib\inst{1} \and
Ruizhong Liu\equalcontrib\inst{2} \and
Yichao Liu\inst{3} \and
Cheng Fan\inst{1} \and
Yu Liu\inst{4} \and
Jianyuan Guo\corrauth\inst{1}
}

\authorrunning{T.~Luo et al.}

\institute{
City University of Hong Kong \and
The Hong Kong University of Science and Technology (Guangzhou) \and
Nankai University \and
Peking University
}

\maketitle
\begingroup
\renewcommand{\thefootnote}{}
\footnotetext{\textsuperscript{*} Equal contribution. \quad
\textsuperscript{\ensuremath{\dagger}} Corresponding author.}
\addtocounter{footnote}{-1}
\endgroup
\begin{figure}[t]
    \centering
    \includegraphics[width=0.95\linewidth]{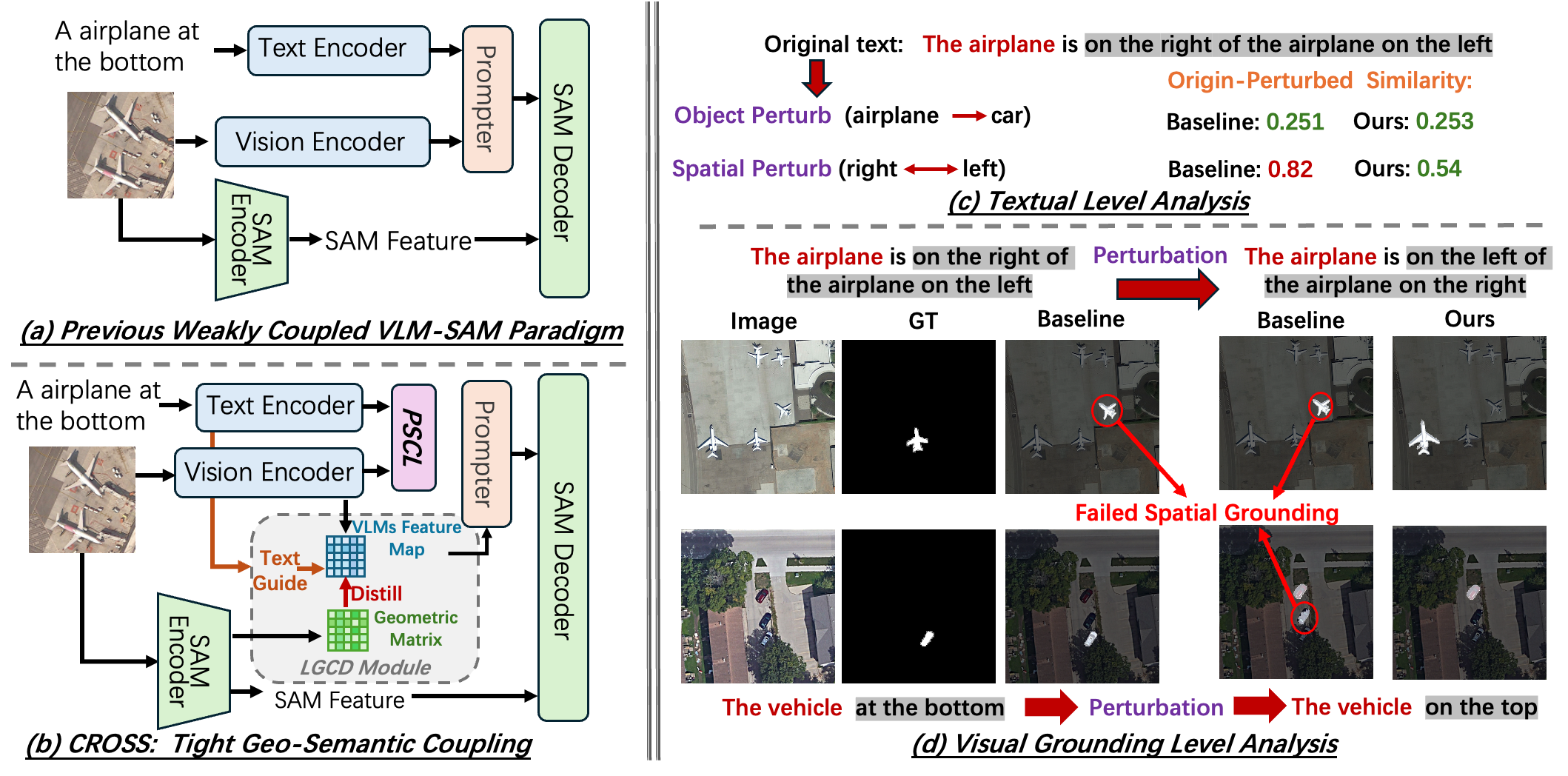}
    \caption{Motivation of the CROSS framework.  \textbf{(Left) Architectural Paradigm Comparison:}  (a) Previous weakly-coupled pipelines treat VLM and SAM as isolated, fragmented modules, where SAM merely serves as a passive executor of explicit prompts. \textit{(b)} Our deeply-coupled CROSS performs text-guided distillation of SAM-derived spatial affinity matrices into the VLM to enforce structural constraints.
\textbf{(Right) Spatial Logic Probing:} \textit{(c)} Textual Level Analysis: Baseline methods show sensitivity to object categories but remain invariant to spatial perturbations, indicating a lack of directional awareness. \textit{(d)} Visual Grounding Analysis: Baseline exhibits logical collapse under complex spatial cues, whereas CROSS maintains logical soundness by precisely anchoring masks to the complete linguistic context. }
    \label{fig:motivation}
\end{figure}
\begin{abstract}

Referring Remote Sensing Image Segmentation (RRSIS) has achieved significant progress through the integration of VLMs and the Segment Anything Model (SAM). However, this progress largely relies on strong pre-trained capabilities, while leaving two fundamental limitations insufficiently addressed: \emph{(1) Architectural Weak-Coupling}, where the unidirectional flow forces reliance on coarse VLM prompts and wastes SAM's pixel-level structural guidance, causing localization drift; and \emph{(2) Object-Centric Semantic Bias}, where models overemphasize dominant object semantics while remaining insensitive to spatial reasoning crucial for RRSIS.
Motivated by these observations, we propose CROSS, a tightly integrated paradigm for RRSIS. First, we introduce Linguistic-Guided Cascaded Distillation (LGCD) to bridge the architectural gap, which distills SAM's geometric affinities as soft regularizers into VLM intermediate layers, injecting dense structural priors to refine localization. Second, Perspective-Spatial Contrastive Learning (PSCL) imposes cross-anchored constraints by mining mask-filtered deceptive distractors and spatial-linguistic counterfactuals as hard negatives, explicitly shattering semantic shortcuts to enforce genuine logical consistency. Extensive experiments on  RRSIS benchmarks demonstrate that CROSS achieves state-of-the-art performance and maintains precise localization even under severe spatial description perturbations, standing as a robust new paradigm for RRSIS. \url{https://clarence-cv.github.io/CROSS/}.


\end{abstract}

\section{Introduction}
\label{sec:intro}

Recent advances in visual understanding\cite{cho2026perceptionlm,lin2025uniworld,li2025stitchfusion,li2026exploring} have increasingly emphasized models’ ability to handle complex semantics, diverse visual patterns\cite{stroke,dig-face}, and open-ended real-world scenarios \cite{luo2024contextuality,li2026towards,xu2026relational,yang2026assignment}. 
Referring Remote Sensing Image Segmentation (RRSIS) \cite{liu2024rotated, yuan2024rrsis,sun2026crobim,li2026open} aims to segment specific targets in Earth observation imagery, strictly adhering to linguistic referring expressions. Unlike natural scene Referring Image Segmentation (RIS) \cite{wang2022cris,liu2017recurrent,huang2025densely}, RRSIS confronts unique complexities: immense visual scale, extreme homogeneity, and dense, multi-scale targets. To tackle these challenges, leveraging foundation models has become a promising paradigm. Within this evolving landscape, research efforts have bifurcated into two primary directions. While Large Multimodal Models (LMMs) \cite{lai2024lisa, liu2023visual,li2025segearth,shabbir2025geopixel,ou2025geopix} excel in holistic understanding and complex reasoning, the specific task of RRSIS demands precise pixel-level localization rather than generative text capabilities. Consequently, the discriminative CLIP-based paradigm (\eg, SigLIP \cite{zhai2023sigmoid,tschannen2025siglip} coupled with SAM \cite{kirillov2023segment,ravi2024sam}) has emerged as the compelling  choice, prioritizing dense feature alignment over semantic reasoning. Adhering to this latter paradigm, pioneering works such as RSRefSeg \cite{chen2025rsrefseg} and  RSRefSeg 2\cite{chen2025rsrefseg2} have sought to transpose this efficient VLM-SAM\footnote{The term \emph{VLM} in our work refers to CLIP-style models. Generative MLLM such as LLaVA~\cite{liu2023visual} or Qwen-VL~\cite{bai2025qwen3} are not considered here, as preliminary evidence suggests they remain less competitive than CLIP-based approaches for precise RRSIS tasks.} pipeline to remote sensing scenarios.

However, rather than achieving a genuine multi-modal synergy tailored for the dense and complex nature of Remote Sensing (RS) scenarios, current adaptations largely coast on inherent pre-trained capabilities of the standalone foundation models. By deeply dissecting this superficial integration, we identify two fundamental bottlenecks that critically paralyze existing VLM-SAM pipelines (as illustrated in Fig. \ref{fig:motivation}).




\noindent$\bullet$ \textbf{Bottleneck I: Architectural Weak-Coupling.}
We characterize existing pipelines as \emph{"weakly coupled"} due to their strictly unidirectional information flow (Fig.~\ref{fig:motivation} a). In such designs, the VLM produces coarse semantic prompts that are subsequently consumed by SAM, while SAM's powerful image encoder remains entirely unused during prompt generation. This design overlooks a critical opportunity: SAM encodes rich pixel-level structural priors that could otherwise guide semantic localization. As a result, VLM-derived dense prompts often exhibit spatially diffuse responses in complex remote sensing scenes, reflecting the limited localization capability of CLIP-style vision encoders when operating without structural constraints. This observation raises a key question: 
\textbf{\textit{Can SAM's structural priors be injected back into the VLM to explicitly constrain semantic localization?}}

\noindent$\bullet$ \textbf{Bottleneck II: Object-Centric Semantic Bias.} 
Beyond architectural isolation, existing pipelines also inherit a strong bias from VLM pre-training. CLIP-style models are primarily trained for global image–text alignment, which encourages strong associations with dominant object semantics while providing limited supervision for spatial relations.
As illustrated in Fig.~\ref{fig:motivation}, our perturbation analysis reveals a critical failure mode. When spatial attributes in the referring expression are altered, the baseline model still produces a high similarity response (Fig.~\ref{fig:motivation}c), yet the predicted localization becomes entirely incorrect (Fig.~\ref{fig:motivation}d). This phenomenon indicates that the model largely relies on object-level semantic cues while remaining insensitive to spatial descriptions that are crucial for RRSIS.
Consequently, a second key question arises: 
\textbf{\textit{How can we explicitly guide the network to overcome this object-centric bias and internalize robust, text-driven spatial reasoning?}}

To directly address these bottlenecks, we propose CROSS: \textbf{C}ascaded Distillation and Dual-Constraint Grounding for \textbf{R}em\textbf{O}te \textbf{S}ensing Referring \textbf{S}egmentation (overall architecture illustrated in Fig. \ref{fig:framework}).  Unlike conventional weakly coupled pipelines, our approach establishes a tightly integrated paradigm utilizing text-guided SAM distillation (Fig. \ref{fig:motivation} b). Specifically, to operationalize this paradigm and overcome the aforementioned architectural isolation, we propose \textbf{L}inguistic-\textbf{G}uided \textbf{C}ascaded \textbf{D}istillation (\textbf{LGCD}). Guided by an adaptive text-driven soft mask, LGCD abstracts SAM's structural priors into geometric affinity matrices and cascadedly distills them into SigLIP's shallow, deep, and final layers. Crucially, transferring relative affinities acts as a soft regularizer. By strategically leveraging different layers, this mechanism explicitly augments spatial configurations without corrupting SigLIP's inherent hierarchical semantic space. This structural enhancement is also visually corroborated by the highly concentrated spatial activations demonstrated in Fig.~\ref{fig:dense_prompt_heatmap}.

Furthermore, to mitigate the object-centric bias inherited from VLM pre-training and strengthen spatial reasoning, we introduce the \textbf{P}erspective-\textbf{S}patial \textbf{C}ontrastive \textbf{L}earning (\textbf{PSCL}) scheme. Formulated as a complementary cross-anchored contrastive space, PSCL imposes dual-level constraints. At the visual perspective level, we utilize the target mask to physically blind the true instance, explicitly mining deceptive background distractors that exhibit high semantic affinity to the text yet violate the intended spatial constraints. Concurrently, at the linguistic spatial level, we synthesize counterfactual text perturbations (\eg, swapping "left" with "right") to strictly penalize spatial inconsistencies. By jointly addressing visual ambiguity and relational reasoning, PSCL encourages robust spatial-semantic grounding, ensuring accurate target localization even under severe linguistic perturbations (Fig. \ref{fig:motivation} right).

Our contributions can be summarized as follows: \textcolor{orange}{\textbf{(i)}} \textbf{Conceptually}, we identify two fundamental bottlenecks in existing RRSIS paradigms: \textit{Architectural Weak-Coupling}, which leads to localization drift due to the unidirectional flow, and \textit{Object-Centric Semantic Bias}, where pre-trained VLMs predominantly rely on dominant object semantics as a shortcut, remaining insensitive to the spatial descriptions crucial for dense localization.
\textcolor{orange}{\textbf{(ii)}}  \textbf{Methodologically}, CROSS introduces two core mechanisms: LGCD, which distills SAM's   geometric affinities as soft structural regularizers to bridge the VLM's geometric-semantic gap; and PSCL, a heterogeneous contrastive construct that rectifies perspective and spatial-linguistic inconsistencies via dual-level hard negative mining. \textcolor{orange}{\textbf{(iii)}} \textbf{Experimentally}, CROSS delivers superior performance and exceptional spatial referring robustness on RRSIS benchmarks.

\section{Related Work}

\noindent\textbf{Referring Remote Sensing Image Segmentation (RRSIS).} As an evolving paradigm in Earth observation \cite{wang2025sopseg, wang2025pcp,zhang2025multi,guan2025sampling,ma2025novel,liu2024crossmatch}, RRSIS enables the extraction of specific spatial regions guided by textual prompts \cite{wang2025rs}. Unlike the remote sensing visual grounding (RSVG) \cite{sun2022visual, zhan2023rsvg, lan2024language} task that focuses on region understanding, RRSIS emphasizes fine-grained pixel-level analysis.  Nevertheless, research in this field is still in its early stages and exploration is limited. 
Yuan et al. \cite{yuan2024rrsis}  first introduced this task,  constructed the RefsegRS dataset and adopted the LAVT\cite{yang2022lavt} framework to solve it. Following this, the first large-scale benchmark, RRSIS-D, was built upon the DIOR-RSVG dataset \cite{zhan2023rsvg} by Liu et al. \cite{liu2024rotated}. To handle arbitrary target orientations and dramatic scale changes, they developed RMSIN, a framework centered on rotational convolutions. In addition, Lei et al. propose FIANet \cite{lei2024exploring}, which focuses more on adaptive understanding of objects at different scales and fine-grained vision-language interaction. Recently, Chen et al. introduced the RSRefSeg\cite{chen2025rsrefseg} and RSRefSeg2\cite{chen2025rsrefseg2} methods by leveraging SigLIP to generate prompts for SAM.

\noindent\textbf{Segment Anything Model.} SAM \cite{kirillov2023segment} and its successor SAM 2 \cite{ravi2024sam} have established a new paradigm for promptable segmentation, demonstrating robust generalization across diverse domains including remote sensing\cite{osco2023segment,yan2023ringmo,gao2025combining}, video tracking, and medical imaging \cite{wang2023samrs, cheng2023segment, yue2024surgicalsam}. While various optimized versions \cite{xiong2024efficientsam, zhong2024convolution} have enhanced its computational performance, a fundamental limitation remains: SAM lacks inherent linguistic understanding. To bridge this gap, MLLM-based frameworks—such as LISA \cite{lai2024lisa}, u-LLaVA \cite{xu2023u}, and EVF-SAM \cite{zhang2024evf}—leverage multimodal large language models to generate text-driven embeddings for SAM. However, these models are primarily optimized for natural images and often struggle with the extreme visual homogeneity and intricate spatial layouts characteristic of remote sensing scenes \cite{zhang2025uniuir}. This underscores the need for specialized structural-semantic alignment tailored for RRSIS.

\noindent\textbf{Vision-Language Models (VLMs)}.  VLMs aim to establish a unified embedding space for cross-modal alignment, supporting a wide range of tasks such as image-text retrieval \cite{radford2021learning,jia2021scaling,zhai2023sigmoid}, reasoning segmentation\cite{lai2024lisa,wang2024llm,liu2025seg,yang2021bottom}, remote sensing analysis\cite{li2023rs,liu2024remoteclip,zhang2024earthgpt,wang2024skyscript,li2025segearth,zhang2026ecrformer,li2025u3m}, and beyond\cite{ito2025feature}. While previous cascaded works \cite{li2024cascade} rely on internal semantic refinement, CROSS leverages SAM to distill external geometric priors, effectively bridging the structural-semantic gap.

\section{Preliminary}
Our referring remote sensing image segmentation framework is built upon a prompt-driven multimodal architecture. The base pipeline consists of three fundamental components: 1) a vision-language foundation model (\textbf{SigLIP 2}\cite{tschannen2025siglip} ) for multimodal feature extraction, 2) a \textbf{Cross-Modal Prompter} that translates semantic features into prompts, and 3) the Segment Anything Model (\textbf{SAM 2}) as the mask decoder.
\subsection{Problem Definition}
Given a remote sensing image $I \in \mathbb{R}^{H_0 \times W_0 \times 3}$ and a natural language referring expression $T$, the remote sensing referring segmentation task aims to generate a binary segmentation mask $M \in \{0,1\}^{H_0 \times W_0}$, where $M_{ij} = 1$ indicates that pixel $(i,j)$ belongs to the target region described by text $T$.

\subsection{Vision-Language Model SigLIP 2}
We employ SigLIP 2 \cite{tschannen2025siglip} to extract dual-modal embeddings. For an image $I$ and text $T$, the encoders $\mathcal{E}_v$ and $\mathcal{E}_t$ generate aligned features:
\begin{equation}
    F_v = \mathcal{E}_v(I) \in \mathbb{R}^{H_v \times W_v \times C}, \quad F_t = \mathcal{E}_t(T) \in \mathbb{R}^{L \times C},
\end{equation}
here $F_v$ represents the  visual features and $F_t$ denotes the textual embeddings of length $L$. Here, $C$ signifies the shared embedding dimension where visual and linguistic semantics are unified through contrastive pre-training.


\subsection{Cross-Modal Prompter}
The prompter receives the multimodal features $F_v$ and $F_t$ as input. It generates a dense semantic prompt $P \in \mathbb{R}^{H_v \times W_v}$ by computing the cross-modal similarity between the visual feature map and the textual embeddings. This prompt $P$ serves as a spatial prior and is passed to the SAM 2 mask decoder to guide target localization and final segmentation.

 \subsection{Segment Anything Model (SAM 2)}SAM 2 \cite{ravi2024sam} functions as the prompt-guided mask decoder. It receives the image features $F_{img}$ from the SAM encoder and the dense semantic prompt $P$. The mask decoder $\mathcal{D}_{mask}$ then integrates these inputs to generate the final binary segmentation mask $M$:
\begin{equation}
M = \mathcal{D}_{mask}(F_{img}, P) \in \{0, 1\}^{H \times W},
\end{equation}

\begin{figure}[t]
    \centering
    \includegraphics[width=1\linewidth]{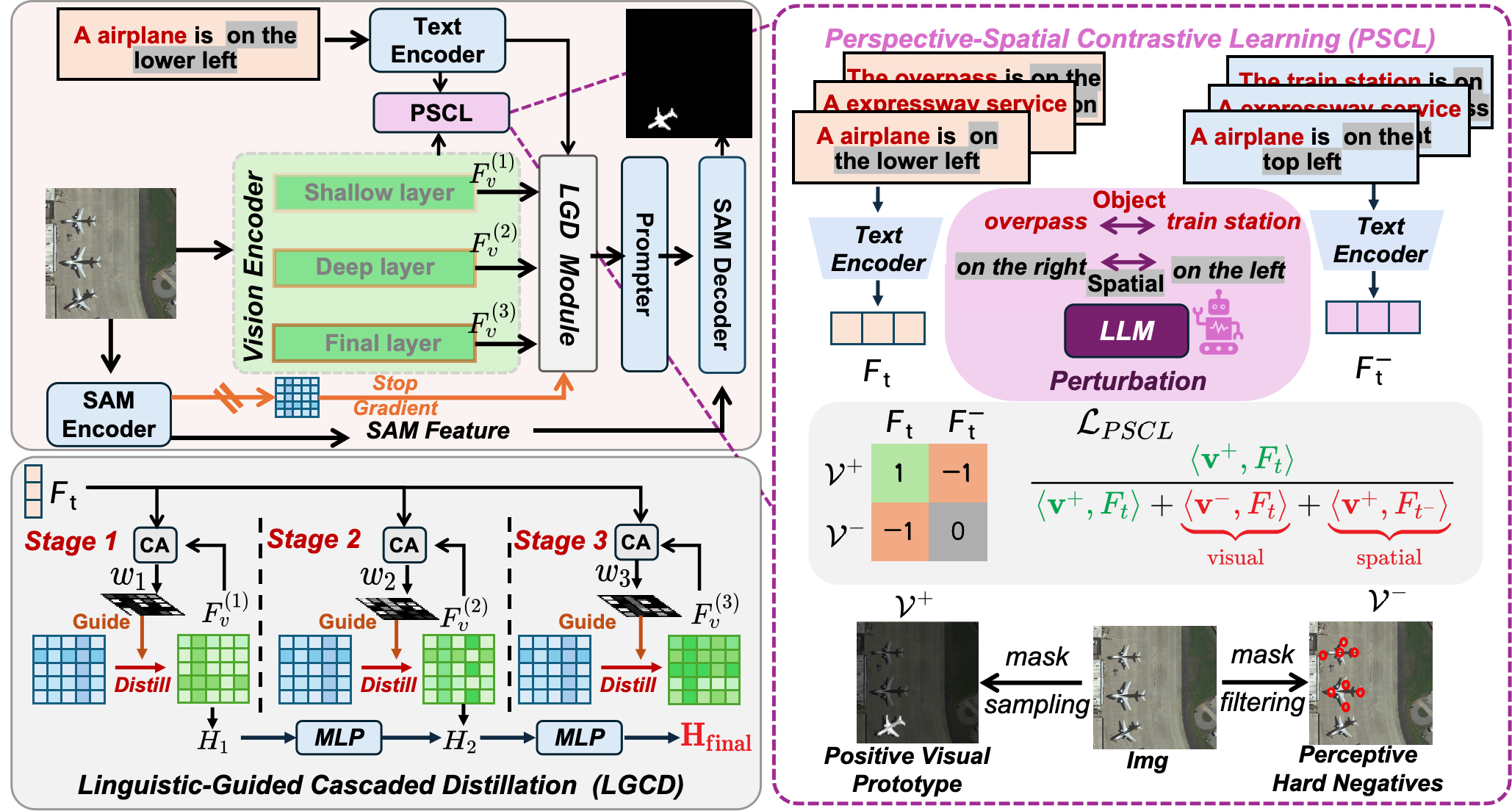}
    \caption{\textbf{Overview of the proposed CROSS.} Our architecture integrates Linguistic-Guided Cascaded Distillation (LGCD) to inject SAM's structural priors into hierarchical VLM layers, and Perspective-Spatial Contrastive Learning (PSCL) which constructs visual and spatial negative samples to enhance spatial-semantic sensitivity.}
    \label{fig:framework}
\end{figure}

 \section{Methodology}
In this section, we present CROSS (Fig. \ref{fig:framework}), which primarily introduces two core mechanisms: Linguistic-Guided Cascaded Distillation (LGCD) and Perspective-Spatial Contrastive Learning (PSCL).

 \subsection{Linguistic-Guided Cascaded Distillation (LGCD)}
To enhance spatial precision, we propose LGCD, which injects class-agnostic geometric priors from   SAM 2 encoder into the SigLIP-2 encoder via a text-aware filtering mechanism.

\noindent\textbf{Cascaded Representation Extraction (CRE).} 
Instead of relying on a singular terminal output, we harvest a cascaded fashion of visual representations $\{F^{(i)}\}_{i=1}^3$ from the shallow, deep, and final blocks of the SigLIP 2 encoder. Each feature map is unrolled into a token sequence $F_v^{(i)} \in \mathbb{R}^{N \times C}$, where $N = H_v \times W_v$ corresponds to the spatial resolution of the vision encoder's patch grid. At each stage $i$, the visual state is updated via a residual injection: $X_i = H_{i-1} + F_v^{(i)}$ ($H_0 = \mathbf{0}$), where the shared spatial dimension $N$ enables direct patch-level summation. To integrate linguistic context as a spatial safeguard, we apply a bidirectional Cross-Attention (CA) mechanism at each stage:

\begin{equation}
H_i = \text{MLP}(\text{LN}(X_i + \gamma_i \cdot \text{CA}(Q{=}X_i, K{=}F_t, V{=}F_t)))
\end{equation}
where $\text{LN}(\cdot)$ and $\text{MLP}(\cdot)$ denote Layer Normalization and a two-layer feed-forward network, and $\gamma_i$ is a zero-initialized learnable scalar to ensure training stability. This process yields the refined feature $H_i \in \mathbb{R}^{N \times C}$.  Simultaneously, we obtain a cross-attention matrix $A_i \in \mathbb{R}^{N \times L}$ derived from the scaled dot-product between visual queries and text keys, which represents the attention weights of $N$ spatial patches over $L$ text tokens.



\noindent\textbf{Text-Guided Relational Distillation (TGD). } At each stage $i$, we distill the topological priors from the  SAM 2 terminal feature  into $H_i$. Given the severe background clutter in remote sensing scenes, a naive dense distillation of SAM's class-agnostic features would introduce massive structural noise. Therefore, we repurpose the cross-attention matrix $A_i$ as a semantic mask to conditionally route only the linguistically relevant topology. For each spatial location $p$, the mask $M_{\text{text}}^i \in \mathbb{R}^{N}$ is aggregated across all $L$ tokens: 

\begin{equation}\label{eq:text_mask}
    M_{\text{text}}^i(p) = \frac{1}{L} \sum_{j=1}^{L} A_{i}^{(p, j)},
\end{equation}
Direct pixel-wise alignment between the heterogeneous latent spaces of SAM and the VLM inevitably causes feature distortion. Instead, we transfer the structural priors by aligning their relative spatial topologies. Specifically, we compute the Gram matrix $\mathcal{G}(\cdot) \in \mathbb{R}^{N \times N}$ to capture the pairwise feature affinities:

\begin{equation}
    \mathcal{G}(F)_{p,q} = \frac{\langle f_p, f_q \rangle}{\|f_p\|_2 \|f_q\|_2},
\end{equation}
where $f_p, f_q$ represent the feature vectors at spatial positions $p$ and $q$. The text-guided relational distillation loss is explicitly formulated to conditionally enforce this structural alignment:
\begin{equation}
    \mathcal{L}_{\text{distill}}^i = \frac{1}{|\Omega|} \sum_{(p,q) \in \Omega} w_{p,q}^i \cdot \big\| \mathcal{G}(H_i)_{p,q} - \mathcal{G}(S)_{p,q} \big\|_2^2,
\end{equation}
where $S$ denotes the feature map extracted from the SAM 2 encoder,  $\Omega = \{(p,q) \mid p,q \in [1,N]\}$ represents all spatial location pairs, and the text-guided soft weight is defined as $w_{p,q}^i = \alpha + (1-\alpha) \cdot (M_{\text{text}}^i(p) + M_{\text{text}}^i(q)) / 2$. The margin $\alpha=0.1$ is used to maintain basic spatial consistency to prevent complete collapse of the background topology. Unlike mask-level distillation, LGCD does not treat SAM 2 as a segmentation oracle or copy its masks as hard pseudo-labels; instead, it transfers text-filtered patch-to-patch affinities as soft relational regularization. The total distillation loss is dynamically aggregated across all cascade stages to provide continuous depth supervision:
\begin{equation}
    \mathcal{L}_{\text{distill}} = \frac{1}{K} \sum_{k=1}^{K} \mathcal{L}_{\text{distill}}^{(k)},
\end{equation}
where $K$ denotes the total number of cascaded stages.


\subsection{Perspective-Spatial Contrastive Learning (PSCL)}
As discussed earlier, pre-trained VLMs inherently suffer from object-centric semantic bias. They tend to take a lazy shortcut, focusing solely on the \textbf{primary subject} (\eg, "vehicle") while ignoring critical \textbf{spatial modifiers} (\eg, "left"). To explicitly shatter this spurious correlation, we propose the PSCL paradigm, which establishes a complementary, cross-anchored contrastive space. Specifically, we dismantle this shortcut from two perspectives:


\noindent\textbf{Perceptive Hard Negatives.} 
We utilize the GT mask to filter out the target region, strictly isolating the background $\Omega_{bg}$. Within $\Omega_{bg}$, pixels exhibiting the highest similarities to the text $F_t$ are considered deceptive distractors, as they likely match the primary subject but violate spatial constraints. Therefore, we construct the hard negative set $\mathcal{V}^- = \{ \mathbf{v}^- \in \Omega_{bg} \mid \text{Top-}K \langle \mathbf{v}^-, F_t \rangle \}$ by dynamically mining these top-$K$ embeddings.


\noindent\textbf{Spatial Counterfactual Negatives.} 
Beyond visual discrimination, we aim to explicitly endow the model with topological reasoning capabilities. To achieve this without intractable geometric modeling, we synthesize spatial counterfactual texts $t^-$. For absolute positional descriptions, we invert the spatial indicators (\eg, "top" $\to$ "bottom"). Crucially, for intricate relative relationships, we systematically swap the subject and the object (\eg, "A golf field is on the left of the green airport" $\to$ "The green airport is on the left of a golf field") to generate logical contradictions. By leveraging a lightweight LLM (\eg, Qwen2.5-7B-Instruct in our experiments) for offline preprocessing, we generate these perturbations to force the model to decode syntactic structures, thereby mitigating the reliance on superficial keyword correlations.

\noindent\textbf{Dual-Constraint Objective.} 
We integrate these perspective and spatial constraints into a unified asymmetric InfoNCE objective, as illustrated in Fig.~\ref{fig:framework}. Given the positive visual prototype $\mathbf{v}^+$ (aggregated from the target mask prediction) and the original text $t$, the PSCL loss is formulated as:
\begin{equation}
\footnotesize
    \mathcal{L}_{PSCL} = - \log \frac{\exp(\langle \mathbf{v}^+, F_t \rangle / \tau)}{ \exp(\langle \mathbf{v}^+, F_t \rangle / \tau) + \sum_{\mathbf{v}^- \in \mathcal{V}^-} \exp(\langle \mathbf{v}^-, F_t \rangle / \tau) + \eta \exp(\langle \mathbf{v}^+, F_{t^-} \rangle / \tau)},
\end{equation}
where $\tau$ is the temperature hyperparameter, $\langle \cdot, \cdot \rangle$ denotes cosine similarity, and $\eta$ is a scaling coefficient balancing the gradient contribution of the spatial logic constraint. Following common practice in previous contrastive learning~\cite{wang2021understanding,radford2021learning}, we set $\tau=0.07$. The scaling factor is set to $\eta=2$ by default.

\subsection{Overall Training Objective}
After the cascaded refinement in the LGCD module, the terminal visual feature $H_{final}$ (i.e., the output of the last cascade stage) and the linguistic embedding $F_t$ are used to generate prompts. Subsequently, these prompts, alongside the image features extracted from the SAM 2 encoder, are fed into the SAM mask decoder to predict the segmentation mask $\hat{M} \in \mathbb{R}^{H \times W}$. Given the corresponding ground truth mask $Y \in \{0, 1\}^{H \times W}$, the overall training objective of CROSS is:
\begin{equation}
    \mathcal{L}_{total} = \lambda_{ce} \mathcal{L}_{\text{ce}}(\hat{M}, Y) + \lambda_{dice} \mathcal{L}_{\text{dice}}(\hat{M}, Y) + \lambda_{1} \mathcal{L}_{\text{distill}} + \lambda_{2} \mathcal{L}_{PSCL},
\end{equation}
where $\mathcal{L}_{\text{ce}}$ and $\mathcal{L}_{\text{dice}}$ denote the cross-entropy loss and DICE loss \cite{milletari2016v}, respectively. $\mathcal{L}_{\text{distill}}$ and $\mathcal{L}_{PSCL}$ represent the cascaded distillation loss and the perspective-spatial contrastive learning loss, respectively.

\section{Experiments}
\subsection{Experimental Settings}
\noindent\textbf{Datasets.} We conduct  comprehensive experiments on two standard RRSIS benchmarks: As the pioneering dataset in this domain, RefSegRS consists of 512$\times$512 resolution images, divided into 2,172, 413, and 1,817 samples for training, validation, and testing, respectively. To further assess scalability, we utilize RRSIS-D, a large-scale benchmark containing 12,181 training, 1,740 validation, and 3,481 test samples, with a higher resolution of 800$\times$800.
 

\noindent\textbf{Evaluation Metrics}
Consistent with prior work in referring image segmentation \cite{yuan2024rrsis, liu2024rotated}, we evaluate our method using three standard metrics: cumulative Intersection over Union (cIoU), generalized Intersection over Union (gIoU), and Precision at specific IoU thresholds (Pr@$X$, where $X \in \{0.5, 0.6, \dots, 0.9\}$). 

\begin{table}[h]
\centering
\caption{Performance comparison across various evaluation metrics on the RefSegRS test dataset.} \label{tab:comparisons-RefSegRS}
\resizebox{0.9\linewidth}{!}{
\begin{tabular}{ c c| *{5}{c} | c c}
\toprule
\textbf{Method} & \textbf{Publication} &\textbf{Pr@0.5} & \textbf{Pr@0.6} & \textbf{Pr@0.7} & \textbf{Pr@0.8} & \textbf{Pr@0.9} & \textbf{cIoU} & \textbf{gIoU} \\ 
\midrule
BRINet \cite{hu2020bi} & CVPR'20 & 20.72 & 14.26 & 9.87 & 2.98 & 1.14 & 58.22 & 31.51 \\
LSCM \cite{hui2020linguistic} & ECCV'20 & 31.54 & 20.41 & 9.51 & 5.29 & 0.84 & 61.27 & 35.54 \\
CMPC \cite{huang2020referring}  & CVPR'20 & 32.36 & 14.14 & 6.55 & 1.76 & 0.22 & 55.39 & 40.63 \\
CMPC+ \cite{liu2021cross} & TPAMI'21 & 49.19 & 28.31 & 15.31 & 8.12 & 2.55 & 66.53 & 43.65 \\
CRIS \cite{wang2022cris} & CVPR'22 & 35.77 & 24.11 & 14.36 & 6.38 & 1.21 & 65.87 & 43.26 \\
LAVT \cite{yang2022lavt} & CVPR'22 & 51.84 & 30.27 & 17.34 & 9.52 & 2.09 & 71.86 & 47.40 \\
CARIS \cite{liu2023caris} & ACM MM'23 & 45.40 & 27.19 & 15.08 & 8.87 & 1.98 & 69.74 & 42.66 \\
RIS-DMMI \cite{hu2023beyond} & CVPR'23 & 63.89 & 44.30 & 19.81 & 6.49 & 1.00 & 68.58 & 52.15 \\
CrossVLT \cite{cho2023cross} & TMM'23 & 71.16 & 58.28 & 34.51 & 16.35 & 5.06 & 77.44 & 58.84 \\
LGCE \cite{yuan2024rrsis} & TGRS'24 & 73.75 & 61.14 & 39.46 & 16.02 & 5.45 & 76.81 & 59.96 \\
DANet \cite{pan2024rethinking} & ACM MM'24&  76.61 & 64.59 & 42.72 & 18.29 &8.04 &  79.53& 62.14 \\
RMSIN \cite{liu2024rotated} & CVPR'24 & 79.20 & 65.99 & 42.98 & 16.51 & 3.25 & 75.72 & 62.58 \\
FIANet \cite{lei2024exploring} & TGRS'24 & 84.09 & 77.05 & 61.86 & 33.41 & 7.10 & 78.32 & 68.67 \\
SBANet \cite{li2025scale} & ISPRS'25 &77.02 & - & 44.15 & - & 8.97 & 79.86 & 62.73 \\
SegEarth-R1 \cite{li2025segearth} & Arxiv'25 &  86.30 & 79.53 &69.57 & 48.87 & 10.73 &79.00 & 72.45 \\
RS2-SAM 2 \cite{rong2025customized} & AAAI'26 & 84.31 & 79.42 & 70.89 & 55.70 & 21.19 & 80.87 & 73.90 \\
RSRefSeg-2\cite{chen2025rsrefseg2} & TGRS'26 &    \ul{88.22}   &   \ul{82.99}  &   \ul{73.97}  &   \ul{60.92}  &   \ul{34.40} &   \ul{81.24}  &   \ul{77.39}  \\
\midrule
  Ours&  - &    \textbf{88.61} &   \textbf{83.98} &   \textbf{76.39} &  \textbf{64.61} &  \textbf{41.32} &  \textbf{83.25} &  \textbf{79.51} \\
\bottomrule
\end{tabular}
}
\end{table}

\subsection{Implementation Details}
CROSS integrates the robust cross-modal alignment of SigLIP 2\cite{tschannen2025siglip} with the high-fidelity segmentation capabilities of SAM 2 \cite{ravi2024sam} into a unified framework.
The SAM 2 variant employed is `sam2.1-hiera-large'\footnote{https://huggingface.co/facebook/sam2.1-hiera-large}, and the SigLIP 2 variant utilized is `siglip2-so400m-patch16-512'\footnote{https://huggingface.co/google/siglip2-so400m-patch16-512}. Following the input requirements of SAM 2 and SigLIP 2, training images were resized to $512 \times 512$ and $1024 \times 1024$, respectively, without any data augmentation. Consistent with the PEFT settings in RSRefSeg 2\cite{chen2025rsrefseg2}, we apply LoRA ($r=16$) to the encoders of both SigLIP 2 and SAM 2.  The trainable parameters consist of the newly introduced low-rank modules, the LGCD module, and the SAM decoder, while all other foundation model backbone weights remain frozen. This configuration results in only \textbf{7.2\%} of the total parameters being updated. For our cascaded distillation, the soft mask weight is set to $\alpha=0.1$. During the contrastive process, we utilize $K=8$ visual negative samples with a spatial penalty weight of $\eta=2$.  The balancing coefficients for the objective function are assigned as $\lambda_{ce} = \lambda_{dice} = 5.0$ following \cite{chen2025rsrefseg2}, while the weights for our auxiliary components are set to $\lambda_{1} = 0.5$ and $\lambda_{2} = 0.2$ via empirical hyperparameter tuning. Training is performed using the AdamW optimizer with a peak learning rate of $1\times10^{-4}$ and a batch size of 8 over 300 epochs. We utilize BF16 precision and the DeepSpeed ZeRO-2 framework on 8 NVIDIA RTX PRO 6000 GPUs.

\begin{table*}[!tbp]
\centering
\caption{Performance comparison across various evaluation metrics on the RRSIS-D test dataset.} \label{tab:comparisons-RRSIS-D}
\resizebox{0.9\linewidth}{!}{
\begin{tabular}{ c c| *{5}{c} | c c}
\toprule
\textbf{Method} & \textbf{Publication} &\textbf{Pr@0.5} & \textbf{Pr@0.6} & \textbf{Pr@0.7} & \textbf{Pr@0.8} & \textbf{Pr@0.9} & \textbf{cIoU} & \textbf{gIoU} \\ 
\midrule
BRINet \cite{hu2020bi} & CVPR'20 & 56.90 & 48.77 & 39.12 & 27.03 & 8.73 & 69.88 & 49.65 \\
CMPC+ \cite{liu2021cross} & TPAMI'21 & 57.65 & 47.51 & 36.97 & 24.33 & 7.78 & 68.64 & 50.24 \\
LAVT \cite{yang2022lavt} & CVPR'22 & 69.52 & 63.63 & 53.29 & 41.60 & 24.94 & 77.19 & 61.04 \\
RIS-DMMI \cite{hu2023beyond}& CVPR'23 & 68.74 & 60.96 & 50.33 & 38.38 & 21.63 & 76.20 & 60.12 \\
CrossVLT \cite{cho2023cross}  & TMM'23 & 70.38 & 63.83 & 52.86 & 42.11 & 25.02 & 76.32 & 61.00 \\
LGCE \cite{yuan2024rrsis} & TGRS'24 & 67.65 & 61.53 & 51.45 & 39.62 & 23.33 & 76.34 & 59.37 \\
EVF-SAM \cite{zhang2024evf} & Arxiv'24 & 72.16 & 66.50 & 56.59 & 43.92 & 25.48 & 76.77 & 62.75 \\
FIANet \cite{lei2024exploring} & TGRS'24 & 74.46 & 66.96 & 56.31 & 42.83 & 24.13 & 76.91 & 64.01 \\
RMSIN \cite{liu2024rotated} & CVPR'24 & 74.26 & 67.25 & 55.93 & 42.55 & 24.53 & 77.79 & 64.20 \\
CADFormer \cite{liu2025cadformer} & JSTARS'25 & 74.20 &67.62& 55.59 &42.37& 23.59 &77.26 &63.77  \\
LSCF \cite{ma2025lscf} & TGRS'25&  74.30 & 67.69  &56.32 &43.08  &25.67  &77.42 & 64.25 \\
RSRefSeg-l \cite{chen2025rsrefseg}	& IGARSS'25 &	74.49&	68.33&	58.73&	48.50&	30.80 &	77.24&	64.67 \\
SegEarth-R1 \cite{li2025segearth} & Arxiv'25 & 76.96  & - & - & -& -& 78.01  & 66.40 \\
RS2-SAM 2 \cite{rong2025customized} & AAAI’26 & 77.56 & 72.34 & 61.76 & 47.92 & 29.73 & 78.99 & 66.72 \\
RSRefSeg-2\cite{chen2025rsrefseg} & TGRS’26 & \ul{80.23} & \textbf{75.78} & \textbf{65.41} & \ul{50.65} & \ul{31.05} & \ul{79.45} & \textbf{69.17} \\
 \midrule
Ours  & - & \textbf{81.24} & \ul{74.56} & \ul{64.80} & \textbf{51.74} & \textbf{32.82} & \textbf{79.89} & \ul{68.92} \\
\bottomrule
\end{tabular}
}
\end{table*}

\subsection{Comparison with State-of-the-Art Methods}
\noindent\textbf{RefSegRS Dataset.}  As detailed in Tab. \ref{tab:comparisons-RefSegRS}, our method establishes a new state-of-the-art across all evaluation metrics, achieving a \textbf{cIoU of 83.25\%} and a \textbf{gIoU of 79.51\%}. Notably, our framework consistently outperforms the strongest competitor, RSRefSeg 2, across the entire precision spectrum. While maintaining a steady lead at looser overlap requirements (\textbf{+0.39\%} at Pr@0.5 and \textbf{+0.99\%} at Pr@0.6), the performance gap widens substantially under more rigorous criteria. Specifically, our method surpasses RSRefSeg 2 by \textbf{+3.69\% at Pr@0.8} and a remarkable \textbf{+6.92\% at Pr@0.9}. This performance trajectory demonstrates that our cascaded distillation and contrastive learning effectively mitigate representational drift, enabling the generation of high-fidelity masks that strictly adhere to target boundaries even under the most stringent overlap constraints.

\noindent\textbf{RRSIS-D Dataset.} Tab. \ref{tab:comparisons-RRSIS-D} demonstrates that CROSS achieves state-of-the-art performance on RRSIS-D, leading in cIoU (\textbf{79.89\%}) and Pr@0.5 (\textbf{81.24\%}). Notably, our framework maintains substantial gains under the most stringent evaluation criteria, outperforming RSRefSeg 2 by \textbf{+1.77\%} at the Pr@0.9 threshold. This performance delta at high precision intervals underscores the efficacy of our Filter-Refine-Verify paradigm; while weakly-coupled baselines suffer from logical drifting, CROSS ensures logical soundness by anchoring masks to the complete linguistic context via LGCD. The marginal gIoU deficit is likely due to the extreme scale diversity of dense small objects in RRSIS-D, where our model prioritizes global logical grounding and high-precision localization over heuristic boundary alignment.

\begin{table*}[t]
  \centering
  \captionsetup{font=small, labelfont=bf}
  \caption{\textbf{Ablation Studies on RefSegRS and RRSIS-D.} (a) Analysis of the proposed \textbf{LGCD} (CRE, TGD) and \textbf{PSCL} ($\mathcal{L}_{\text{PSCL}}$) modules. (b) Impact of different negative sample types in PSCL. All results in cIoU (\%).}
  \label{tab:ablations}
  

  \begin{subtable}{0.53\textwidth} 
    \centering
    \resizebox{\textwidth}{!}{ 
      \begin{tabular}{ccc|cc}
      \toprule
      \multicolumn{2}{c}{\textbf{LGCD Module}} & \multicolumn{1}{c|}{\textbf{PSCL}} & \multicolumn{2}{c}{\textbf{Benchmarks}} \\
      \cmidrule(lr){1-2} \cmidrule(lr){3-3} \cmidrule(lr){4-5}
      \textbf{CRE} & \textbf{TGD} & \textbf{$\mathcal{L}_{\text{PSCL}}$} & \textbf{RefSegRS} & \textbf{RRSIS-D} \\
      \midrule
                 \xmark    &      \xmark          &    \xmark               & 80.82          & 77.80          \\
      \checkmark  &      \xmark          &    \xmark               & 82.55          & 78.63          \\
      \checkmark  & \checkmark  &        \xmark           & 83.15          & 79.02          \\
              \xmark       &      \xmark          & \checkmark     & 82.11          & 78.12          \\    
      \rowcolor{gray!10}
      \checkmark  & \checkmark  & \checkmark     & \textbf{83.25}  & \textbf{79.89}  \\
      \bottomrule
      \end{tabular}
    }
    \caption{\textbf{Main Components}}
    \label{tab:ablation_main}
  \end{subtable}
  \hfill 
  \begin{subtable}{0.45\textwidth} 
    \centering
    \resizebox{\textwidth}{!}{ 
      \begin{tabular}{cc|cc}
      \toprule
      \multicolumn{2}{c|}{\textbf{PSCL Negatives}} & \multirow{2}{*}{\textbf{RefSegRS}} & \multirow{2}{*}{\textbf{RRSIS-D}} \\
      \cmidrule{1-2}
      \textbf{Perspective} & \textbf{Spatial} &  &  \\
      \midrule
      \xmark      & \xmark           & 83.15          & 79.02          \\
      \checkmark      & \xmark           & 83.21          & 79.34          \\
      \xmark          & \checkmark       & 83.20          & 78.70          \\
      \rowcolor{gray!10}
      \checkmark      & \checkmark       & \textbf{83.25} & \textbf{79.89} \\
      \bottomrule
      \end{tabular}
    }
    \caption{\textbf{Formulation of $\mathcal{L}_\text{PSCL}$}}
    \label{tab:ablation_loss}
  \end{subtable}
\end{table*}

 \begin{figure}[t]
  \centering
  \begin{subfigure}{0.24\linewidth}
    \centering
    \includegraphics[width=\linewidth]{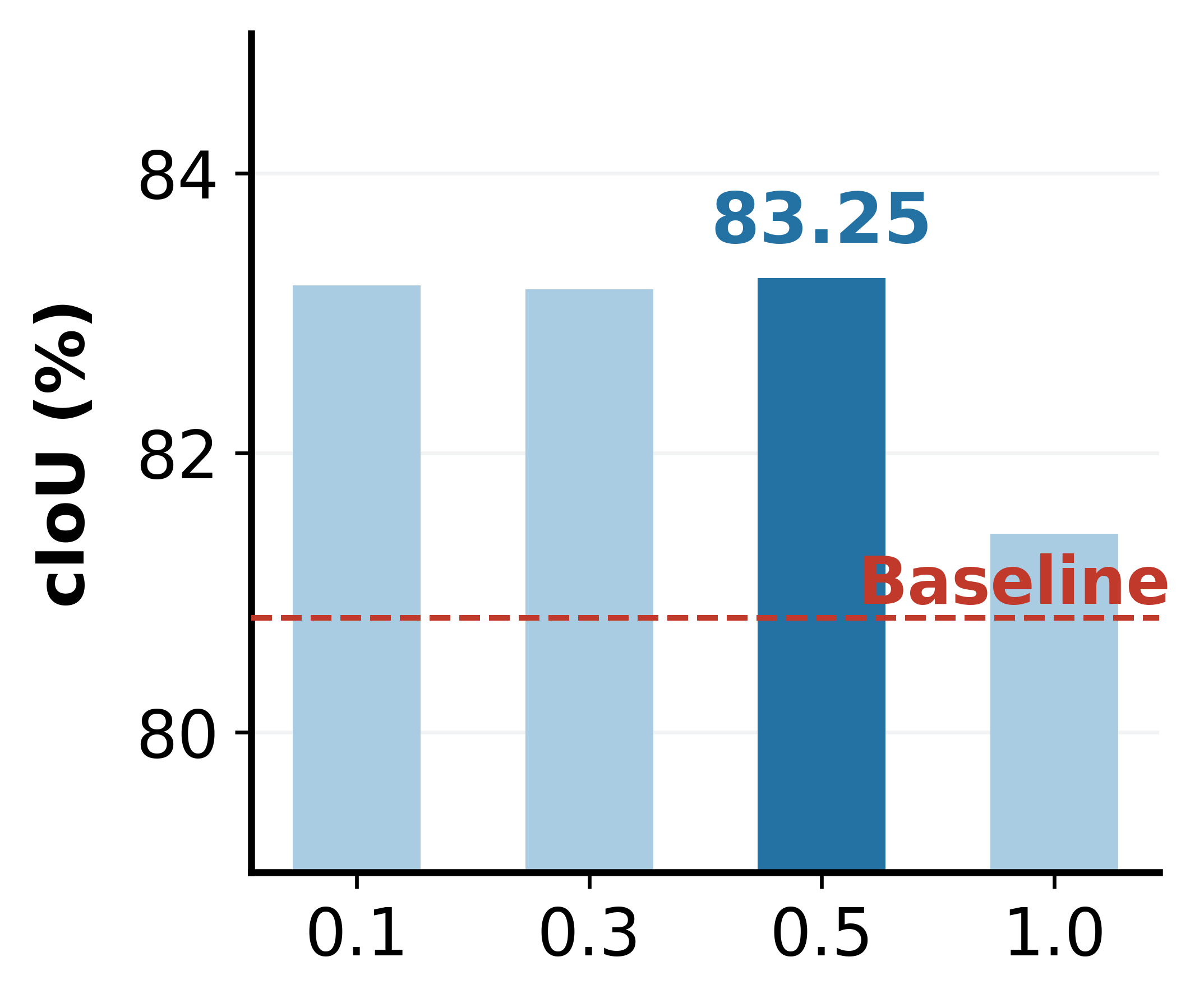}
    \caption{Weight $\lambda_1$}
  \end{subfigure}
  \hfill
  \begin{subfigure}{0.24\linewidth}
    \centering
    \includegraphics[width=\linewidth]{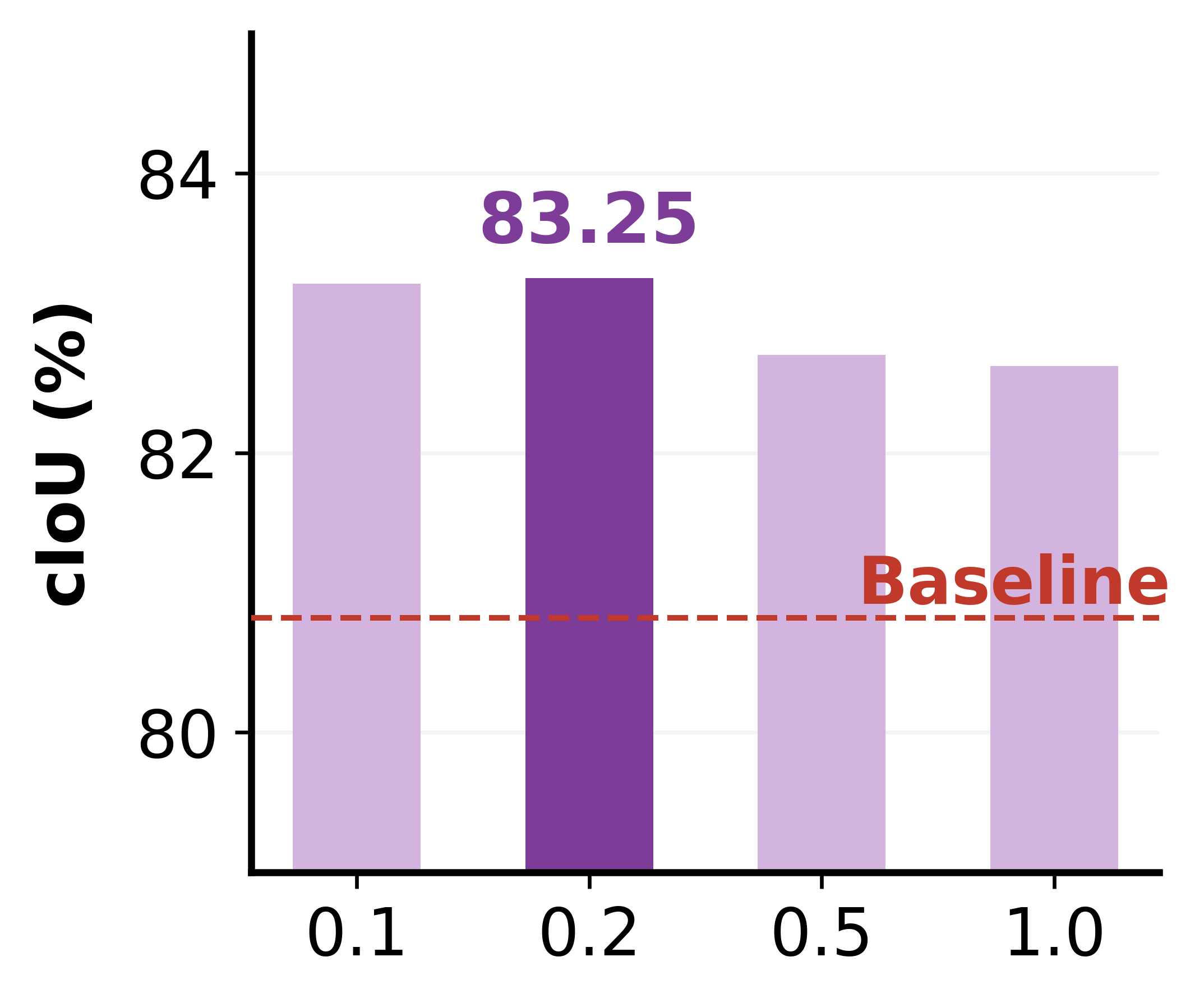}
    \caption{Weight $\lambda_2$}
  \end{subfigure}
  \hfill
  \begin{subfigure}{0.24\linewidth}
    \centering
    \includegraphics[width=\linewidth]{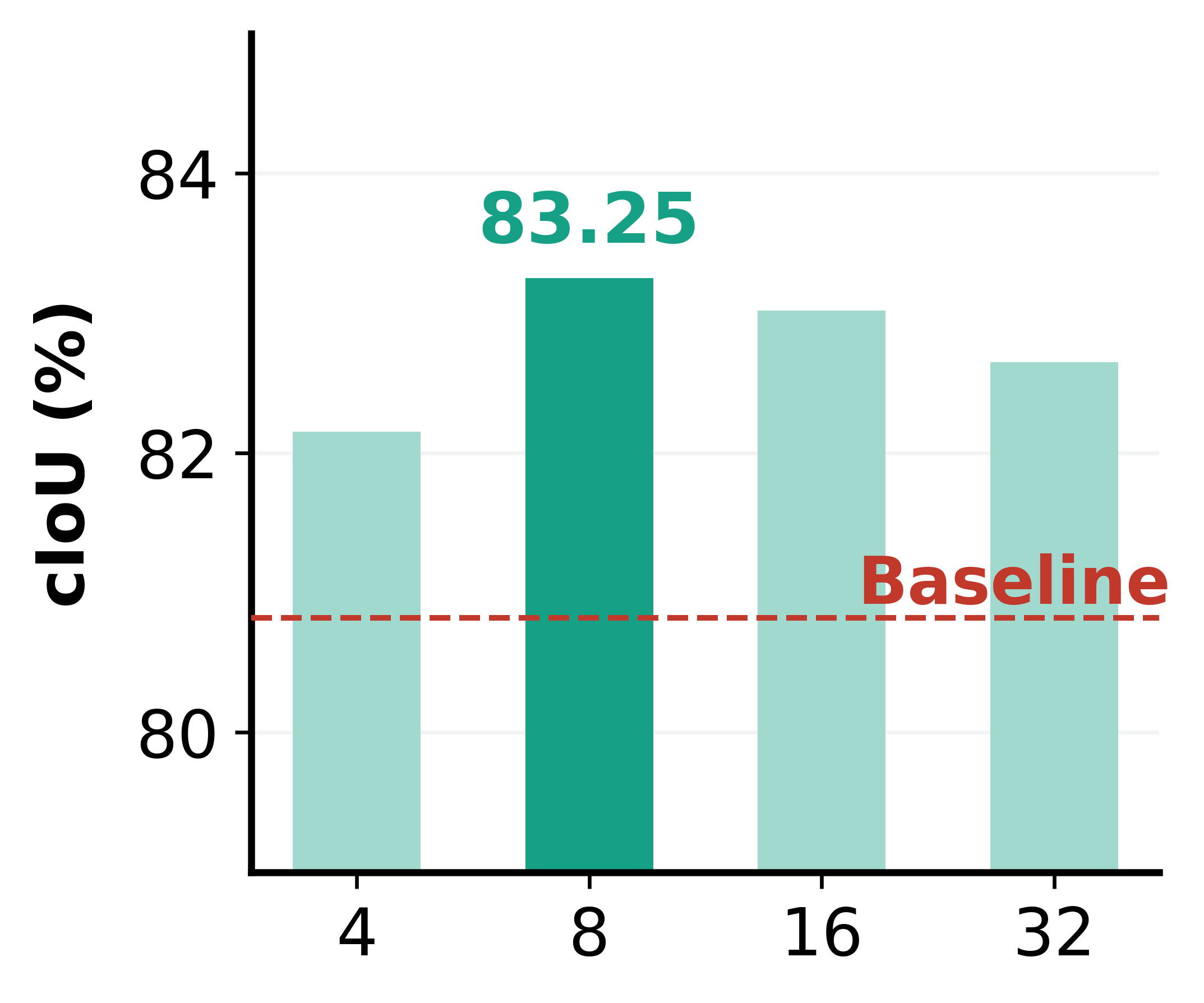}
    \caption{Candidates $K$}
  \end{subfigure}
  \hfill
  \begin{subfigure}{0.24\linewidth}
    \centering
    \includegraphics[width=\linewidth]{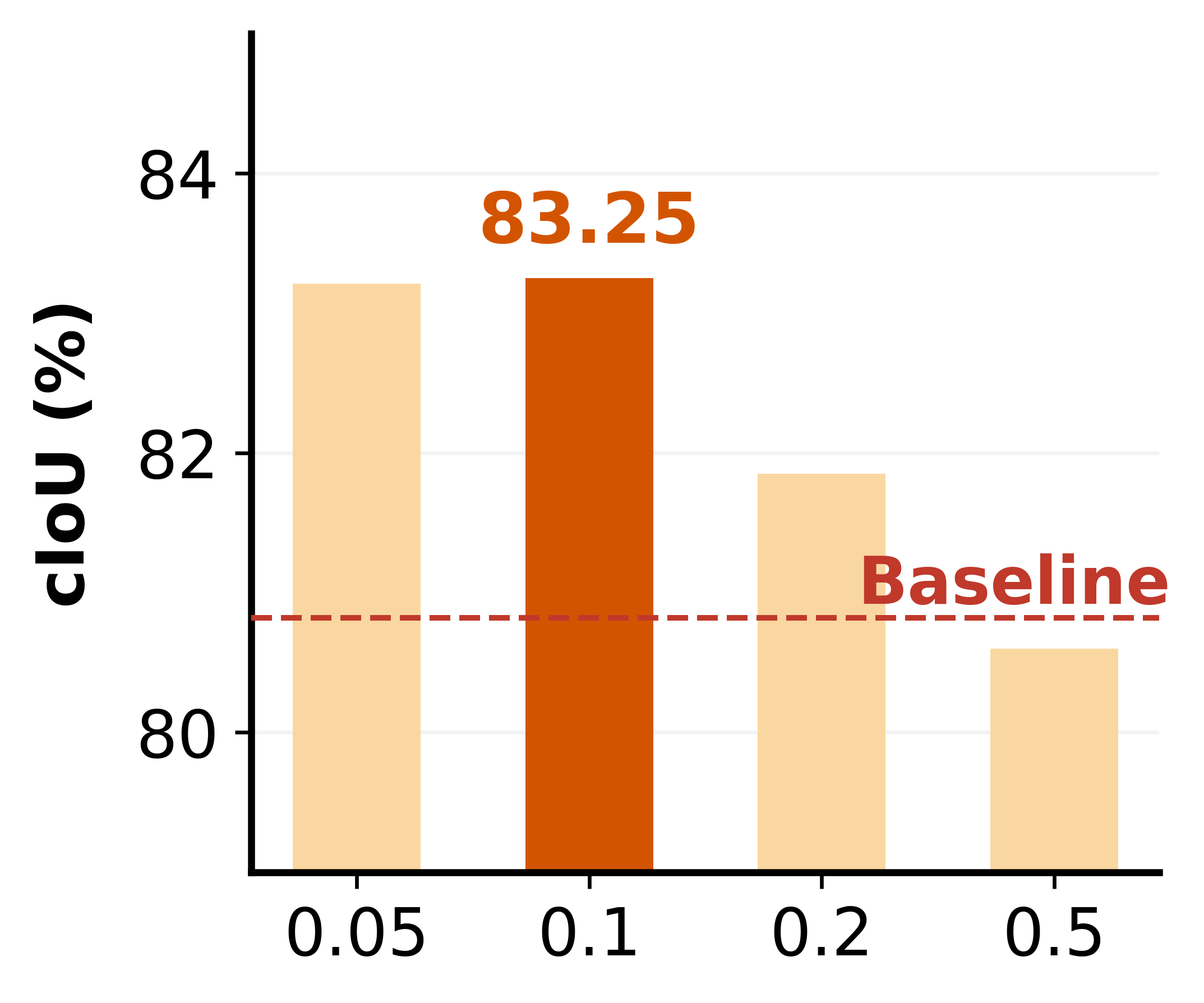}
    \caption{Weight $\alpha$}
    
  \end{subfigure}

\caption{\textbf{Sensitivity analysis on RefSegRS with respect to cIoU.} We evaluate the impact of \textbf{(a)} $\lambda_1$, \textbf{(b)} $\lambda_2$, \textbf{(c)} top-$k$ candidates $K$, and \textbf{(d)} soft weight $\alpha$. The results demonstrate that CROSS maintains stable and superior performance across a broad range of hyperparameter settings.}
  \label{fig:hyperparameters}
\end{figure}

\subsection{Ablation Studies}
\noindent\textbf{Effectiveness of Main Components.}
We conduct a series of component-wise ablations on RefSegRS and RRSIS-D to verify the contribution of each module, as reported in Tab. \ref{tab:ablations} (a). The baseline (SigLIP 2 + SAM 2) achieves \textbf{80.82\%} cIoU on RefSegRS.  Individually, we first conduct experiments by integrating Cascaded Representation Extraction (\textbf{CRE}), which yields a significant gain (\textbf{+1.73\%}), validating that transferring SAM's geometric priors is essential for dense prediction. Meanwhile, Text-Guided Relational Distillation (\textbf{TGD})  leverages textual cues to mitigate background clutter within intermediate layers, effectively filtering out irrelevant noise across different stages. Jointly, these modules reach \textbf{83.15\%}, demonstrating that bridging the semantic-geometric gap is vital for robust performance. Finally, the inclusion of $\mathcal{L}_{\text{PSCL}}$ yields the peak result \textbf{83.25\%} on RefSegRS and \textbf{79.89\%} on RRSIS-D, as it provides the necessary discriminative power to suppress illusory binding against visually similar distractors.

\noindent\textbf{Investigation of Negative Samples in Contrastive Object.} We investigate the negative formulations in Tab. \ref{tab:ablations} (b). Relying exclusively on spatial contrast leads to a performance decrease (\textbf{-0.05\%} on RefSegRS), as spatial cues without visual grounding cause ambiguity in cluttered remote sensing scenes. Conversely, incorporating visual negatives improves the result to \textbf{83.21\%}. Ultimately, combining both yields the peak \textbf{83.25\%} on RefSegRS and \textbf{79.89\%} on RRSIS-D, ensuring discriminability in both perspective and spatial dimensions.

\begin{wraptable}{r}{0.55\textwidth}
  \centering
  \small 
  \setlength{\tabcolsep}{4pt} 
  
  \begin{tabular}{c|cc}
    \toprule
    \textbf{Layer Indices} & \textbf{RefSegRS} & \textbf{RRSIS-D} \\
    \midrule
    $\{18, 24, 27\}$ & 83.19 & 79.71 \\
    $\{5, 15, 27\}$  & 82.76 & 79.42 \\
    $\{25, 27, 27\}$ & 83.14 & 79.50 \\
    \rowcolor{gray!15}
    $\{9, 18, 27\}$ & \textbf{83.25} & \textbf{79.89} \\
    \bottomrule
  \end{tabular}
  \caption{Ablation on layer selection (cIoU \%).}
  \label{tab:ablation_layers} 
\end{wraptable}

\noindent\textbf{Hyperparameter Analysis.} 
\textbf{Firstly}, we conduct a comprehensive sensitivity analysis of four key hyperparameters in Fig. \ref{fig:hyperparameters}. Crucially, across comprehensive tested ranges, CROSS consistently maintains a superior performance margin over the baseline (\textbf{80.82\%}), as indicated by the red dashed lines.
The optimal weights are achieved at $\lambda_1 = 0.5$ and $\lambda_2 = 0.2$; increasing $\lambda_1$ beyond 0.5 tends to cause gradient dominance that destabilizes the joint optimization. For the hard negative set, $K=8$ performs best by maintaining a balanced contribution between perceptive and spatial negatives. Regarding the text-guided background soft filtering, $\alpha=0.1$ is found to be ideal, whereas a larger value (\eg, 0.5) leads to excessive suppression of potential target features. These stable trends across both datasets validate our empirical configurations and ensure the model's robustness. \textbf{Secondly,}  we investigate the choice of intermediate layers, adopting $\{9, 18, 27\}$ as the final configuration, as shown in Tab. \ref{tab:ablation_layers}. Notably, we do not emphasize the strict optimality of these specific indices, as our primary goal is simply to ensure the extraction of features from diverse intermediate stages. The marginal performance degradation observed only with excessively shallow layers (\eg, $\{5, 15, 27\}$) confirms that the framework remains robust to layer selection within a reasonable range.

\subsection{Further Analysis and Discussion}

\noindent$\bullet$\textbf{Parameter Overhead Analysis.} As summarized in Tab. \ref{tab:params}, CROSS updates \textbf{106.09 M} parameters, representing a \textbf{7.29\%} trainable ratio. Compared to the baseline and RSRefSeg 2, our framework introduces a parameter increase of \textbf{1.23\%} and \textbf{0.57\%}, respectively, relative to the total 1.4B+ backbone scale. This increase is primarily attributed to the integrated cascaded distillation module. Given the resulting performance gains across multiple evaluation metrics, we consider this additional parameter cost to be acceptable.

\begin{table}[h]
\centering
\caption{Comparative analysis of parameter scales. The total and trainable parameters are evaluated across the baseline (SAM 2 + SigLIP 2), RSRefSeg 2, and our CROSS.}
\label{tab:params}
\scriptsize 
\setlength{\tabcolsep}{4pt} 
\begin{tabular}{l|cc|c}
\toprule
\textbf{Method} & \textbf{Total Params (B)} & \textbf{Trainable Params (M)} & \textbf{Ratio (\%)} \\ \midrule
Baseline (SAM 2+SigLIP 2) & 1.438 B & 88.41 M & 6.15\% \\
RSRefSeg 2  & 1.447 B & 97.87 M & 6.76\% \\
\midrule
\textbf{CROSS} & \textbf{1.455 B} & \textbf{106.09 M} & \textbf{7.29\%} \\ \bottomrule
\end{tabular}
\end{table}

\begin{figure}[h]
    \centering
    \includegraphics[width=1\linewidth]{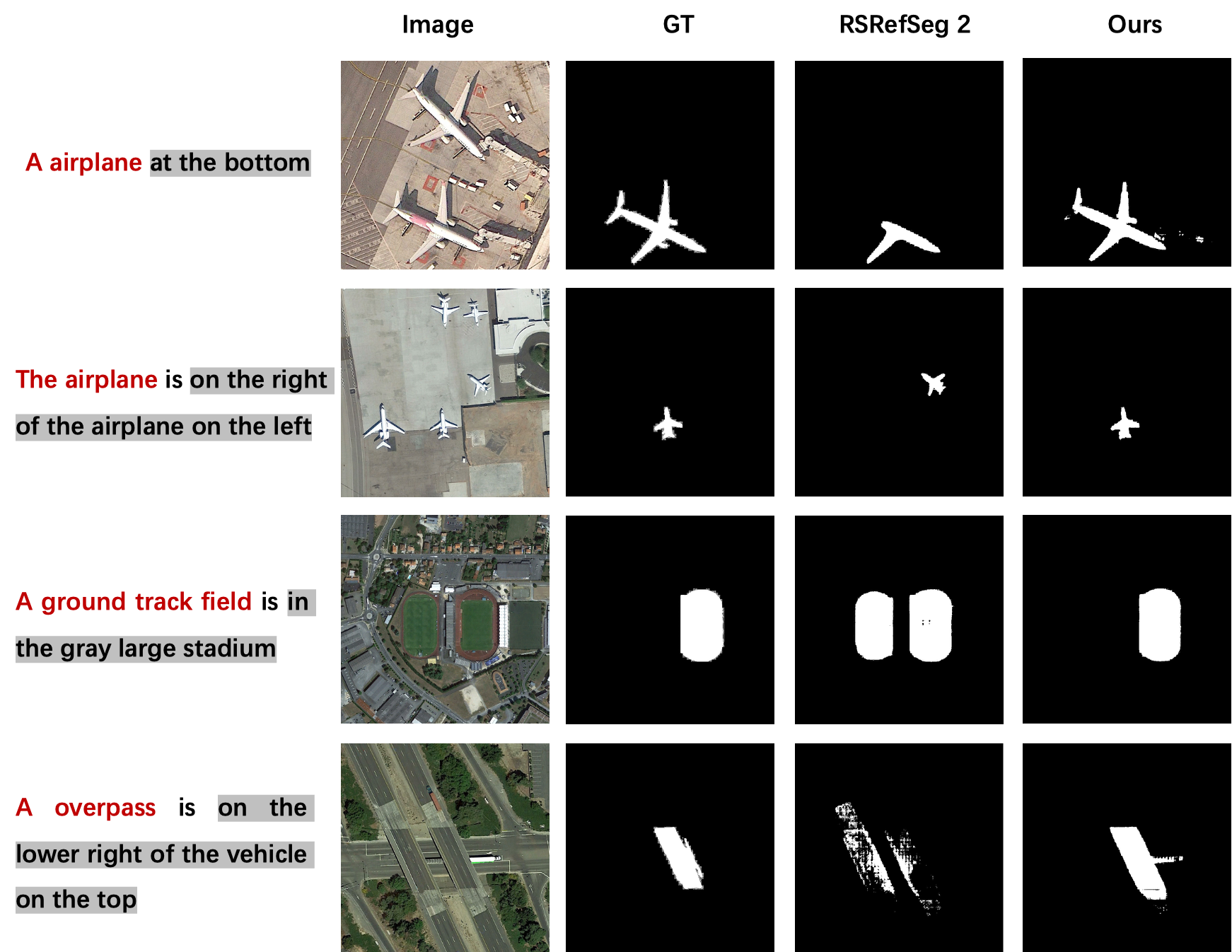}
    \caption{Visualization result on RRSIS-D. The targets and spatial descriptions are highlighted in \textcolor{red!80!black}{\textbf{red}} and \textcolor{black!80}{\textbf{gray}}, respectively. Compared to RSRefSeg 2, our method achieves more accurate spatial referring and precise boundaries.}
    \label{fig:cases}
\end{figure}

\noindent$\bullet$\textbf{Robust Spatial Grounding.} Fig. \ref{fig:cases} probes the grounding logic of CROSS against SOTA method RSRefSeg 2 in complex scenarios. For the nested prompt \textbf{\textit{“\target{The airplane} is \spatial{on the right of the airplane on the left}”}}, RSRefSeg 2 fails to parse the full relational context, drifting toward the partial clause \textbf{\textit{“on the right”}}. In addition, in the scenario \textbf{\textit{“\target{A ground track field} is \spatial{in the gray large stadium}”}}, where two identical track fields are present, the baseline erroneously segments both instances. This oversight indicates a failure to process the spatial qualifier \textbf{\textit{“is in the gray large stadium”}}, effectively reducing the logic-driven task to a simple category-level search.
CROSS resolves these via \textbf{PSCL}, which enforces strict logical alignment through contrastive verification. By penalizing "logical collapse," CROSS effectively filters distractors and ensures  \spatial{spatial constraint} is actively utilized for disambiguation.

\begin{figure}[t] 
    \centering
    
    \begin{minipage}[b]{0.48\linewidth}
        \centering
        \includegraphics[width=\linewidth]{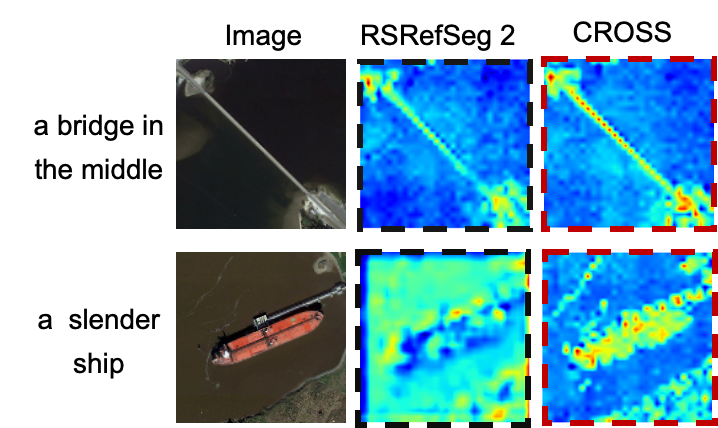}
        \caption{\textbf{Visualization of dense prompt heatmaps.} 
        Compared to the baseline RSRefSeg 2, the heatmaps of CROSS are significantly more concentrated.}
        \label{fig:dense_prompt_heatmap}
    \end{minipage}
    \hfill 
    \begin{minipage}[b]{0.48\linewidth}
        \centering
        \includegraphics[width=\linewidth]{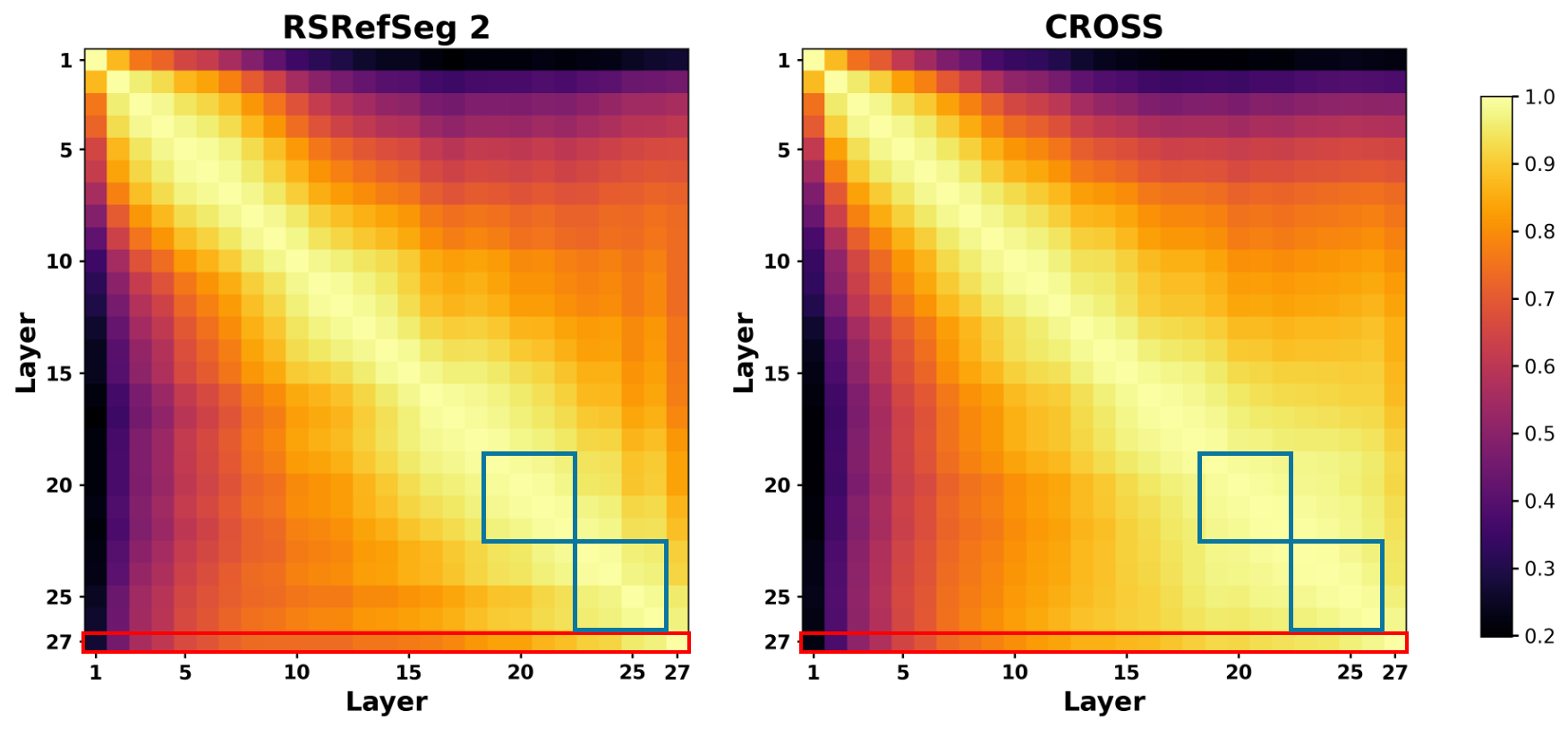}
        \caption{\textbf{Internal representation analysis via CKA \cite{kornblith2019similarity}.} 
        Comparison of layer-wise similarity heatmaps. The \target{red box} highlights stable alignment, while the \textcolor{blue}{blue box} illustrates smoother transitions.}
        \label{fig:cka_analysis}
    \end{minipage}
    
\end{figure}

\noindent$\bullet$\textbf{Effects of Structural Distillation.} As shown in Fig.~\ref{fig:dense_prompt_heatmap}, CROSS yields   more concentrated dense prompt heatmaps compared to RSRefSeg 2. This provides direct evidence that our distillation successfully injects SAM's structural priors into the VLM,   enhancing fine-grained localization precision.
 
\noindent$\bullet$\textbf{Structural Consistency Analysis.} To quantitatively assess the internal representation dynamics, we employ Centered Kernel Alignment (CKA) \cite{kornblith2019similarity} to visualize layer-wise feature similarities. Fig.~\ref{fig:cka_analysis} compares our CROSS with the baseline. The \target{red box} reveals stable alignment between the final layer and preceding deep hierarchies, while the \textcolor{blue}{blue box} illustrates notably smoother transitions between adjacent layers. 
These phenomena stem directly from our cascaded distillation design. By injecting a shared SAM-derived geometric affinity matrix across multiple layers, we establish a consistent structural prior that regularizes the representational trajectory. Crucially, to strictly prevent this shared constraint from inducing detrimental layer homogenization (i.e., collapsing hierarchical feature diversity), the distillation is dynamically modulated by a \textit{layer-specific} soft mask (Eq.~\ref{eq:text_mask}).  By guiding the shared geometric anchor through this layer-adaptive masking, our design ensures that the network internalizes robust structural awareness while strictly preserving its capacity for adaptive feature evolution.


\section{Conclusion}
In this paper, we presented CROSS, a novel framework for Referring Remote Sensing Image Segmentation (RRSIS). Our research identifies and addresses two fundamental bottlenecks in existing foundation-model adaptations: the semantic-geometric gap arising from weakly coupled architectures, and the illusory spatial-semantic binding inherent in pre-trained VLMs. By introducing Linguistic-Guided Cascaded Distillation (LGCD) for structural prior distillation and Perspective-Spatial Contrastive Learning (PSCL) for contrastive refinement, CROSS effectively restores representational smoothness and enhances geometric fidelity. Extensive experiments on RRSIS benchmarks demonstrate that CROSS achieves state-of-the-art performance across comprehensive evaluation metrics, particularly under stringent precision requirements, and demonstrates a superior understanding of spatial referring.

\section*{Acknowledgements}
This work was supported by the Hong Kong RGC under Grant 21218026, and the City University of Hong Kong under Grants 9382010 and 7020171.


%
%
\bibliographystyle{splncs04}
\bibliography{refs}

\clearpage

\clearpage

\setcounter{section}{0}
\section*{Appendix}

\section{ Spatial Counterfactual Generation: Rules, Cases, and LLM Ablation} \label{sec1}
This section provides the implementation details of our spatial counterfactual generation pipeline. We first outline the hierarchical construction rules and the exact prompt templates used for text synthesis. Next, we present specific text-to-text transformation cases to illustrate the generation process. Finally, we report the quantitative ablation results across different LLM engines (Qwen2.5, Llama-3.1, and GPT-4o) on the RRSIS-D dataset.

\subsection{Construction Rules}

\begin{tcolorbox}[
    colback=gray!8, 
    colframe=gray!50, 
    arc=4pt, 
    boxrule=0.5pt, 
    title=\textbf{Construction Rules for Spatial Counterfactual Prompts}, 
    coltitle=black, 
    fonttitle=\bfseries\small, 
    left=6pt, right=6pt, top=6pt, bottom=6pt 
]
\small 
To synthesize spatial counterfactuals for PSCL, we prompt \textbf{Qwen2.5-7B-Instruct} to parse the original referring expressions and apply rule-based perturbations based on their spatial syntax, strictly following these hierarchical rules:

\vspace{0.5em}
\textbf{Rule 1: Absolute Position Perturbation (Single Entity)} \\
\textit{Condition:} The prompt contains only absolute spatial anchors without referencing other objects. \\
\textit{Action:} Invert the absolute directional keywords (\eg, top $\leftrightarrow$ bottom, left $\leftrightarrow$ right). \\
\textit{Example:} ``A baseball field at the \textbf{bottom}'' $\rightarrow$ ``A baseball field at the \textbf{top}''.

\vspace{0.5em}
\textbf{Rule 2: Relative Relation Disruption (Target-Reference Pairs)} \\
\textit{Condition:} The prompt describes a spatial relationship between a target object and a reference object.
\begin{itemize}
    \setlength{\itemsep}{0pt}
    \item \textbf{2a. Heterogeneous Entities (Target $\neq$ Reference):} \\
    \textit{Action:} Swap the semantic nouns of the target and reference objects while strictly freezing the spatial relational words. \\
    \textit{Example:} ``A \textbf{harbor} is on the right of the \textbf{gray and black slender small ship}'' $\rightarrow$ ``A \textbf{gray and black slender small ship} is on the right of the \textbf{harbor}''.
    
    \item \textbf{2b. Homogeneous Entities (Target $=$ Reference):} \\
    \textit{Action:} Perturb the relative spatial relational words while freezing the entity descriptions to penalize relational collapse. \\
    \textit{Example:} ``The tennis court is on the \textbf{right} of the tennis court on the \textbf{left}'' $\rightarrow$ ``The tennis court is on the \textbf{left} of the tennis court on the \textbf{right}''.
\end{itemize}

\vspace{0.5em}
\textbf{Rule 3: No Spatial Description} \\
\textit{Condition:} The prompt contains purely appearance-based descriptions without any spatial or relational priors. \\
\textit{Action:} Retain the original text without perturbation (Excluded from spatial contrastive pairs). \\
\textit{Example:} ``The large green field'' $\rightarrow$ \textit{SKIP (No action)}.
\end{tcolorbox}
%


\subsection{Spatial Counterfactual Cases}
\begin{figure}[t]
    \centering
    \includegraphics[width=1\linewidth]{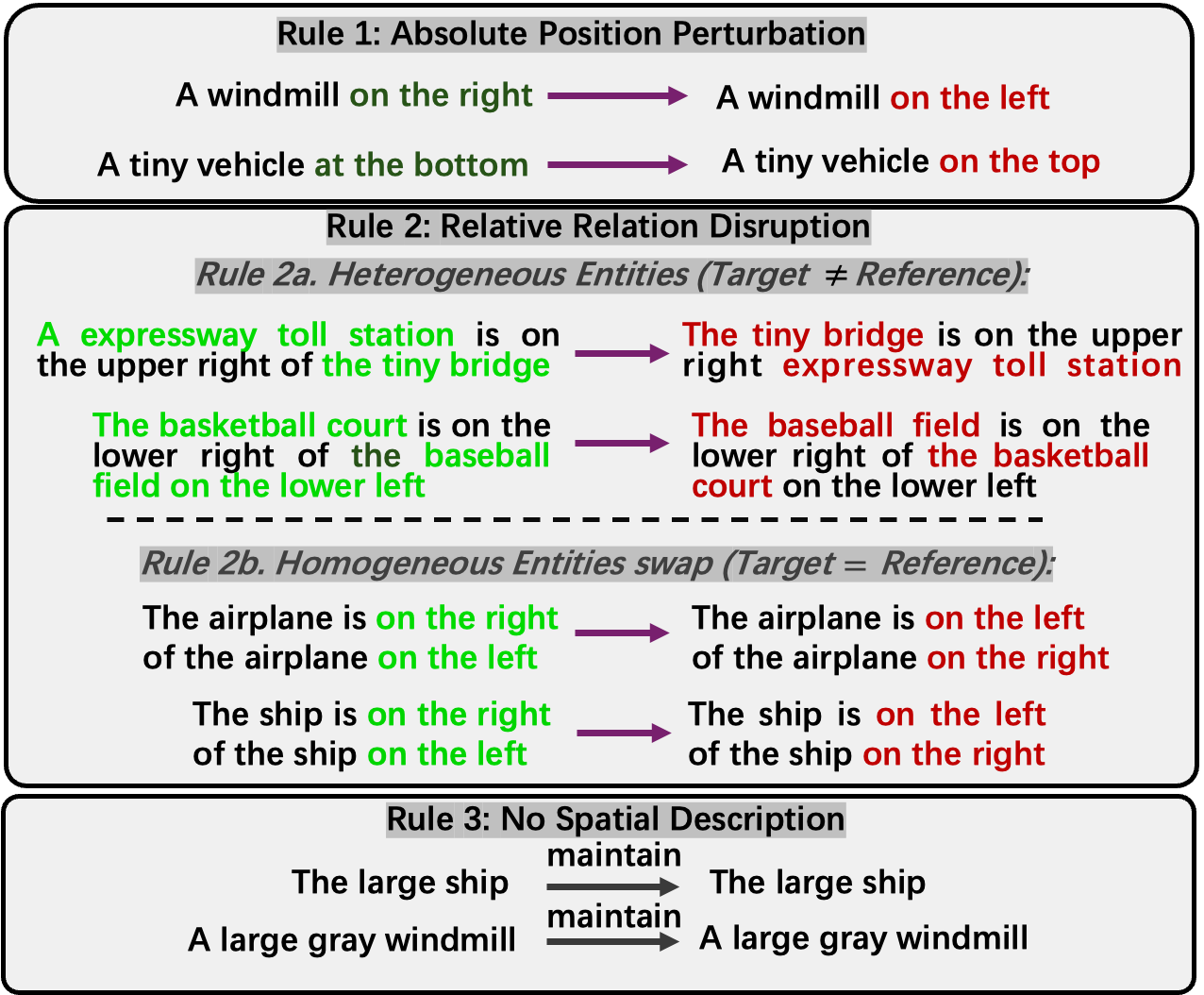}
    \caption{Cases  of spatial counterfactual sample construction.}
    \label{fig:spatial_case}
\end{figure}

Fig. \ref{fig:spatial_case} provides several examples detailing how the original referring expressions are transformed into spatial counterfactuals based on our rules. By explicitly  constructing these hard linguistic negatives, we force the model to look beyond mere appearance priors and genuinely comprehend the underlying spatial logic, thereby significantly enhancing its robustness against deceptive object-centric biases in complex remote sensing scenes.

\subsection{ Robustness to the Choice of LLM Generator}

A natural question arises regarding whether the performance of our Perspective-Spatial Contrastive Learning (PSCL) heavily relies on the specific ecosystem or the emergent capabilities of the Large Language Model (LLM). To investigate this, we conduct an ablation study using a state-of-the-art closed-source model (\textbf{GPT-4o}) and a highly representative open-source baseline (\textbf{Llama-3.1-8B-Instruct}), replacing our default \textbf{Qwen2.5-7B-Instruct} as the syntactic engine. \textbf{We specifically evaluate this on the RRSIS-D test dataset, as it is uniquely characterized by highly complex spatial relationships, making it the ideal testbed for assessing spatial counterfactuals.}

As reported in Tab. \ref{tab:llm_ablation_rrsisd}, using our default Qwen2.5-7B-Instruct as the anchor, the performance across different LLM generators exhibits extremely marginal fluctuations. Notably, no single model strictly dominates across all metrics. For instance, while GPT-4o performs marginally better under strict IoU thresholds (e.g., +0.06\% in Pr@0.9), and Llama-3.1-8B slightly leads in gIoU (+0.04\%), our default Qwen2.5-7B maintains a slight edge in Pr@0.5, Pr@0.8, and cIoU. This mixed yet uniformly excellent performance essentially confirms that the success of CROSS stems from the explicit structural logic of our predefined generation rules, rather than the idiosyncratic bias of a specific LLM. 

\begin{table}[h]
\centering
\caption{Ablation on different LLM generators for spatial counterfactuals on the \textbf{RRSIS-D test dataset}. We set our default Qwen2.5-7B-Instruct as the baseline. The superscripts indicate the absolute performance gap. The highly consistent performance proves that our rule-based generation is robust and not tied to any specific LLM ecosystem.} 
\label{tab:llm_ablation_rrsisd}
\renewcommand{\arraystretch}{1.15} 
\resizebox{\linewidth}{!}{ 
\begin{tabular}{ l | *{5}{c} | c c }
\toprule
\textbf{LLM Generator} & \textbf{Pr@0.5} & \textbf{Pr@0.6} & \textbf{Pr@0.7} & \textbf{Pr@0.8} & \textbf{Pr@0.9} & \textbf{cIoU} & \textbf{gIoU} \\ 
\midrule
\rowcolor{gray!10}
\textbf{Qwen2.5-7B-Instruct (Ours)} & \textbf{81.24} & 74.56 & 64.80 & \textbf{51.74} & 32.82 & \textbf{79.89} & 68.92 \\
\midrule
GPT-4o & 81.21$^{\textcolor{red}{-0.03}}$ & 74.53$^{\textcolor{red}{-0.03}}$ & \textbf{64.86}$^{\textcolor{green!60!black}{+0.06}}$ & 51.71$^{\textcolor{red}{-0.03}}$ & \textbf{32.88}$^{\textcolor{green!60!black}{+0.06}}$ & 79.86$^{\textcolor{red}{-0.03}}$ & 68.89$^{\textcolor{red}{-0.03}}$ \\
Llama-3.1-8B-Instruct & 81.18$^{\textcolor{red}{-0.06}}$ & \textbf{74.60}$^{\textcolor{green!60!black}{+0.04}}$ & 64.76$^{\textcolor{red}{-0.04}}$ & 51.70$^{\textcolor{red}{-0.04}}$ & 32.78$^{\textcolor{red}{-0.04}}$ & 79.84$^{\textcolor{red}{-0.05}}$ & \textbf{68.96}$^{\textcolor{green!60!black}{+0.04}}$ \\
\bottomrule
\end{tabular}
}
\end{table}

\section{Computational Complexity and Profiling of LGCD}\label{sec2}

To achieve fine-grained spatial reasoning, the Linguistic-Guided Cascaded Distillation (LGCD) module aligns visual features with structural priors. However, dense relational distillation inherently introduces potential computational bottlenecks. In this section, we analyze the theoretical complexity and empirical efficiency of our asymmetric architectural design.

\vspace{0.5em}
\noindent\textbf{Feature Resolution Bounding.} 
To ensure computational tractability, LGCD avoids operating on the native high-resolution image space. Instead, the cascading process is strictly confined to the compact feature space of the SigLIP-2 encoder. Specifically, the visual feature maps are extracted at a fixed spatial resolution of $H \times W = 32 \times 32$. This 2D grid is subsequently flattened into a 1D token sequence of length $N = 1024$, with a channel dimension of $C=1152$.

To map the heterogeneous representations, we explicitly \textbf{down-pool} the high-resolution structural features of SAM 2 ($S$) to match the bounded sequence length ($N=1024$) of SigLIP-2. To illustrate the computational necessity of this design, we explicitly recall the relational distillation mechanism from the main text. The Gram matrix $\mathcal{G} \in \mathbb{R}^{N \times N}$ is first computed to capture pairwise feature affinities:
\begin{equation}
    \mathcal{G}(F)_{p,q} = \frac{\langle f_p, f_q \rangle}{\|f_p\|_2 \|f_q\|_2} \quad \text{\footnotesize \textbf{(Eq. 5 in the main paper)}}
\end{equation}
Subsequently, the distillation loss $\mathcal{L}_{\text{distill}}$ enforces structural alignment based on this dense matrix:
\begin{equation}
    \mathcal{L}_{\text{distill}}^i = \frac{1}{|\Omega|} \sum_{(p,q) \in \Omega} w_{p,q}^i \cdot \big\| \mathcal{G}(H_i)_{p,q} - \mathcal{G}(S)_{p,q} \big\|_2^2 \quad \text{\footnotesize \textbf{(Eq. 6 in the main paper)}}
\end{equation}
Because the pairwise affinity computation requires calculating distances for all $(p,q)$ pairs, it inherently triggers a quadratic cost w.r.t the sequence length $N$. 

\vspace{0.5em}
\noindent\textbf{Complexity Analysis and Empirical Profiling.} 
Given the sequence length $N$ and channel dimension $C$, the theoretical time complexity for the Gram matrix computation is strictly bounded at $\mathcal{O}(B \cdot N^2 \cdot C)$, and the space complexity at $\mathcal{O}(B \cdot N^2)$, where $B$ denotes the batch size. Upsampling the SigLIP-2 features to the native image resolution would result in a quadratic explosion in $N^2$, leading to a prohibitive memory footprint.
To quantitatively validate the efficiency of our down-pooling strategy, we profile the computational overhead of the text-guided relational distillation, as detailed in Tab. \ref{tab:gram_overhead}.

\begin{table}[h]
    \centering
    \caption{\textbf{Empirical Profiling of the Text-Guided Relational Distillation.} Metrics are evaluated per forward pass with a batch size of $B=8$. By bounding the sequence length to $N=1024$, the $\mathcal{O}(N^2)$ explicit pairwise affinity computation introduces negligible latency and memory overhead.}
    \label{tab:gram_overhead}
    \vspace{-0.5em}
    \resizebox{0.95\linewidth}{!}{
    \begin{tabular}{l c l}
        \toprule
        \textbf{Configuration / Metric} & \textbf{Empirical Profiling} & \textbf{Theoretical Complexity} \\
        \midrule
        Spatial Resolution ($H \times W$) & $32 \times 32$ & - \\
        Flattened Sequence Length ($N$) & 1,024 & - \\
        Feature Dimension ($C$) & 1,152 & - \\
        \midrule
        Explicit Gram Matrix Size & 32.00 MB & Space: $\mathcal{O}(B \cdot N^2)$ \\
        Computation FLOPs & 9.68 GFLOPs & Time: $\mathcal{O}(B \cdot N^2 \cdot C)$ \\
        Forward Latency Overhead & 0.82 ms & - \\
        \bottomrule
    \end{tabular}
    }
\end{table}

As reported in Tab. \ref{tab:gram_overhead}, maintaining $N=1024$ effectively limits the explicit memory allocation of the dense Gram matrix to merely 32.00 MB. The pairwise alignment computation incurs a minimal overhead of 9.68 GFLOPs, translating to an additional 0.82 ms per forward pass. This empirical evidence demonstrates that our dimensional constraint successfully circumvents the memory scaling issues of dense topological alignment.

\vspace{0.5em}
\noindent\textbf{Asymmetric Inference Paradigm.} 
Finally, the LGCD module follows an asymmetric training-inference paradigm. While storing multi-layer SAM 2 features and intermediate computational graphs naturally increases the peak memory footprint during training, the dense structural alignment strictly serves as a topological regularizer. Consequently, the extraction of SAM 2 features and the Gram matrix computations are entirely discarded during inference. The test-time architecture operates solely as a lightweight feature refiner, preserving the original inference speed (FPS) and deployment efficiency of the baseline framework.

\section{Qualitative Visualizations}\label{sec3}
\begin{figure}[h]
    \centering
    \includegraphics[width=0.94\linewidth]{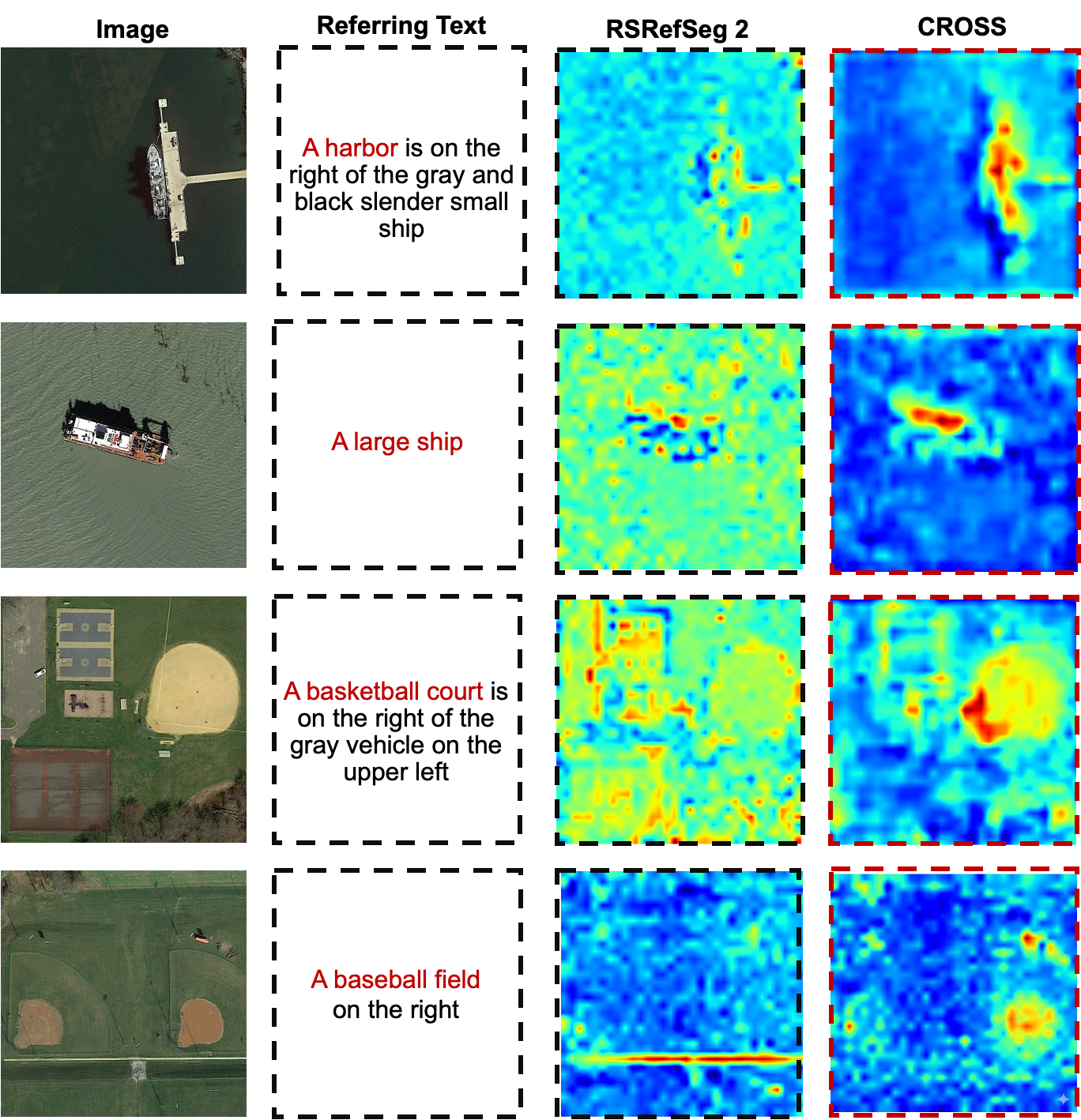}
    \caption{\textbf{Additional Qualitative Visualizations of Dense Prompts.} We provide supplementary comparisons of attention heatmaps (dense prompts) generated by RSRefSeg 2  and our CROSS framework on the RRSIS-D dataset. The visualizations explicitly illustrate that our proposed framework effectively mitigates diffuse attention patterns and corrects chaotic spatial logic, providing highly focused visual prompts for the subsequent SAM 2 decoder.}
\label{fig:dense_prompts_vis}
    \label{fig:placeholder}
\end{figure}

\begin{figure}[h]
    \centering
    \includegraphics[width=1\linewidth]{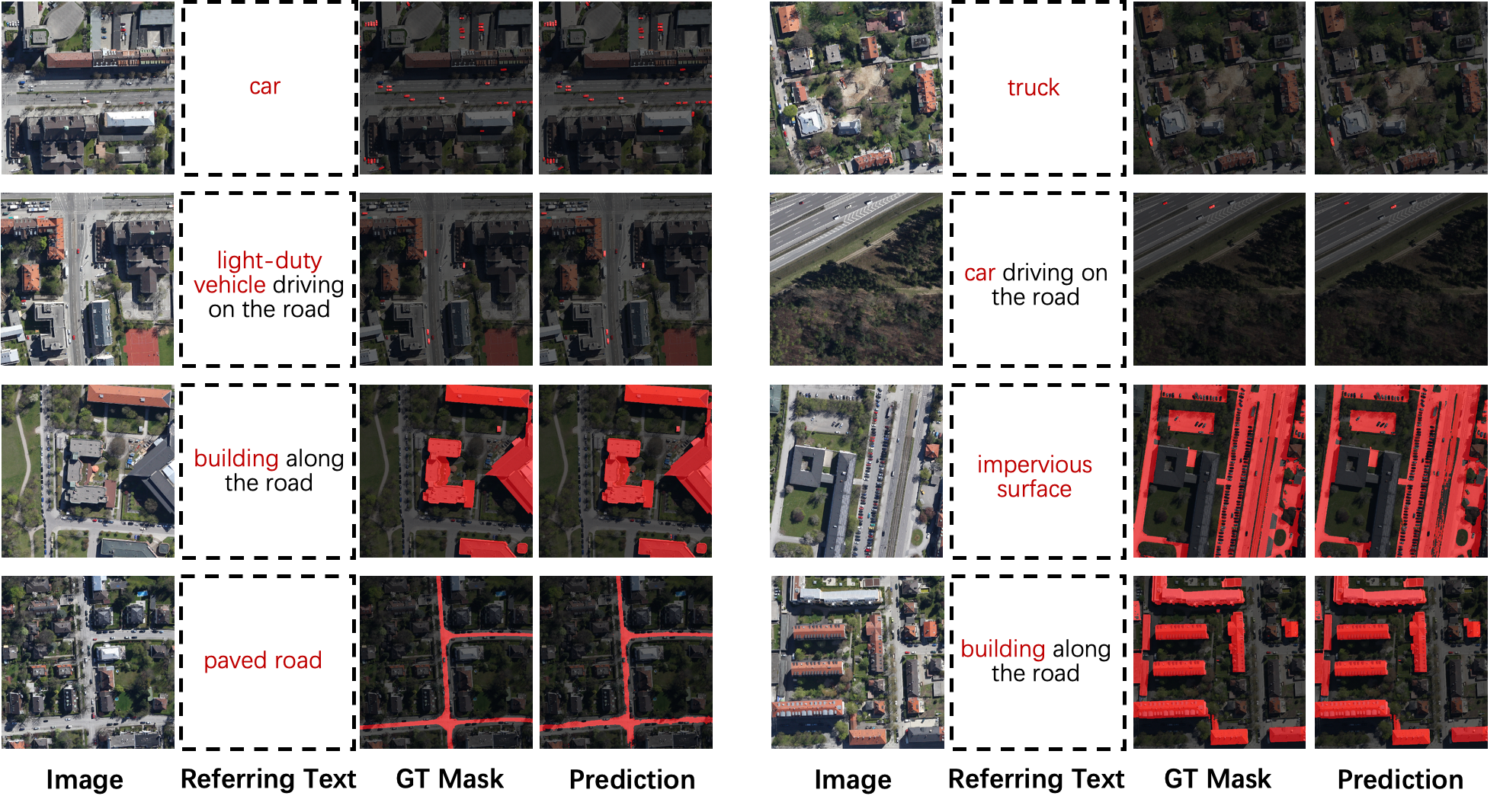}
 \caption{\textbf{Qualitative Segmentation Results on the RefSegRS Dataset.} We present the predicted masks generated by CROSS alongside their corresponding ground truth (GT).}
    \label{fig:refsegrs_vis}
\end{figure}
To further evaluate the fine-grained grounding capabilities of our CROSS framework, we provide supplementary qualitative visualizations in this section. We first investigate the intermediate dense prompts (attention heatmaps) to reveal the spatial reasoning mechanisms. Subsequently, we present the final segmentation masks on the RefSegRS dataset to demonstrate the framework's robustness across diverse object scales (\eg, large regions vs. tiny instances) and complex multi-target scenarios.

\subsection{Visualization of Dense Prompts}
To intuitively demonstrate the effectiveness of our framework, we visualize the dense visual prompts (\eg, attention heatmaps) generated prior to the SAM 2 decoder. As shown in Fig. \ref{fig:dense_prompts_vis}, the proposed LGCD module effectively mitigates diffuse attention patterns, enabling the model to extract highly concentrated structural features that suppress irrelevant background clutter. Furthermore, by integrating Perspective-Spatial Contrastive Learning (PSCL), CROSS exhibits a robust comprehension of directional semantics. For instance, given complex spatial queries such as \textit{``A basketball court is on the right of the gray vehicle on the upper left''} and \textit{``A baseball field on the right''}, our method precisely anchors on the correct target objects. In stark contrast, RSRefSeg 2 yields chaotic and severely misaligned activation maps. We argue that feeding such ambiguous and scattered visual prompts to the SAM 2 decoder is highly unreliable and inevitably leads to segmentation failures. By generating concentrated, target-reliable dense prompts, CROSS guarantees robust mask generation in complex remote sensing scenes.

\subsection{Qualitative Segmentation Results on RefSegRS}

While the visual prompts illustrate intermediate spatial logic, we additionally present final mask predictions on the RefSegRS dataset. This dataset features highly complex scenes with extreme scale variations (\eg, expansive infrastructures vs. tiny vehicles) and dense multi-target clusters, posing significant challenges for precise localization.

As illustrated in Fig. \ref{fig:refsegrs_vis}, our predicted masks exhibit a remarkably high alignment with the ground truth (GT). Despite the presence of multiple distracting instances with similar visual semantics (\eg, adjacent buildings, densely packed vehicles), CROSS consistently delineates target objects with precise boundaries. Regardless of whether the referred target is a massive building or a tiny, isolated instance, the generated masks maintain high fidelity without severe over-segmentation or under-segmentation. These visual results explicitly demonstrate that our dual-constraint grounding mechanism effectively handles severe background interference and multi-target ambiguity, ensuring robust mask generation across arbitrary object scales.

\section{Additional Ablation}\label{sec4}
\begin{figure}
     \centering
     \includegraphics[width=0.4\linewidth]{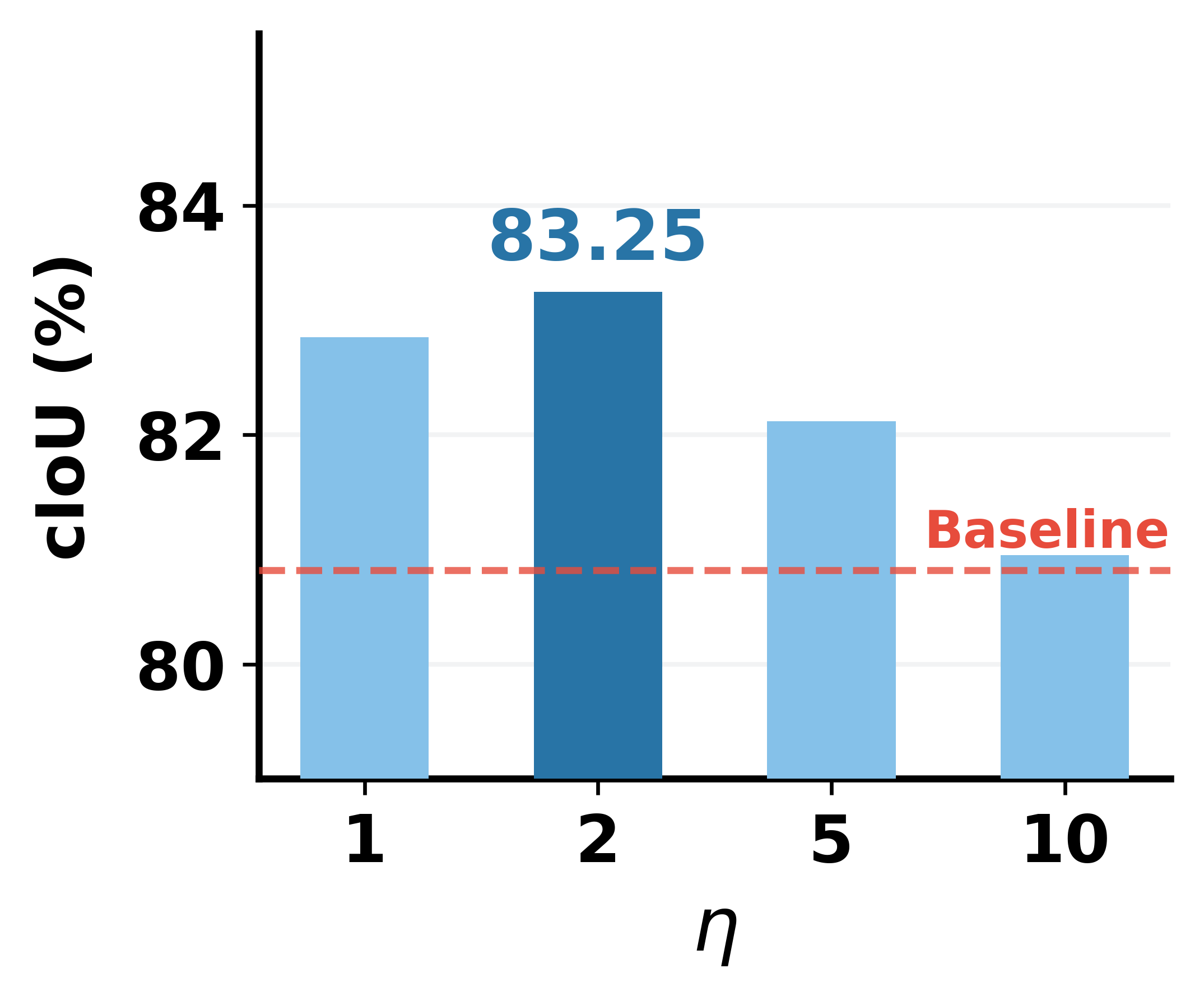}
     \caption{\textbf{Sensitivity Analysis of the Contrastive Hyperparameter $\eta$.} We vary $\eta \in \{1, 2, 5, 10\}$ to observe its impact on the segmentation performance (cIoU) on the RefSegRS dataset. The red dashed line denotes the baseline performance. The optimal balance between spatial reasoning and fine-grained localization is achieved at $\eta = 2$.}
     \label{fig:eta}
 \end{figure}

In our Perspective-Spatial Contrastive Learning (PSCL) module, the hyperparameter $\eta$ explicitly controls the optimization strength of the spatial counterfactual regularization. To justify our empirical setting and evaluate the framework's sensitivity, we conduct a comprehensive hyperparameter ablation on the challenging RefSegRS dataset.  As shown in Fig. \ref{fig:eta}, $\eta=2$ achieves the optimal balance. A smaller weight (\eg, $\eta=1$) under-penalizes spatial errors, failing to suppress distractors. Conversely, an excessively large $\eta$ (\eg, $\eta \ge 5$) over-dominates the optimization, leading to the under-utilization of hard visual negatives.

 \section{Additional Stability Analysis}\label{sec5}

To evaluate training stability, we conduct two additional retraining/evaluation runs on RRSIS-D. As shown in Tab.~\ref{tab:rrsisd_retrain}, CROSS exhibits only normal training fluctuations and consistently preserves strong cIoU and high-precision localization.

\begin{table}[h]
\centering
\setlength{\tabcolsep}{2.6pt}
\renewcommand{\arraystretch}{0.98}
\caption{Additional retraining/evaluation runs on RRSIS-D.}
\begin{tabular}{lccccc|cc}
\hline
Method & Pr@0.5 & Pr@0.6 & Pr@0.7 & Pr@0.8 & Pr@0.9 & cIoU & gIoU \\
\hline
CROSS reported & 81.24 & 74.56 & 64.80 & 51.74 & 32.82 & 79.89 & 68.92 \\
CROSS 2$^{nd}$ run & 81.36 & 75.42 & 65.18 & 51.86 & 32.47 & 79.97 & 69.04 \\
CROSS 3$^{rd}$ run & 81.28 & 74.60 & 64.84 & 51.77 & 32.79 & 79.92 & 68.90 \\
\hline
\end{tabular}
\label{tab:rrsisd_retrain}
\end{table}

\end{document}